\documentclass[letterpaper,twocolumn,10pt]{article}

\usepackage{usenix-2020-09}
\usepackage[]{hyperref}
\usepackage[normalem]{ulem}
\usepackage{graphicx}
\usepackage{algorithm}
\usepackage{algpseudocode}
\usepackage{tikz}
\usetikzlibrary{arrows}
\usepackage{enumerate}
\usepackage{enumitem}
\usepackage{comment}
\usepackage{pifont}
\usepackage{hyperref}

\usepackage{booktabs}
\usepackage{tabularx}
\usepackage{makecell}
\usepackage{amssymb}
\usepackage[table,dvipsnames]{xcolor}
\usepackage{caption}
\usepackage{subcaption}
\usepackage[normalem]{ulem}
\usepackage{siunitx}

\usepackage{algorithm}
\usepackage{algpseudocode}
\usepackage{amsmath}

\usepackage[many]{tcolorbox}    	% for COLORED BOXES (tikz and xcolor included)
\usepackage{multicol}               % for MULTICOLUMNS

\usepackage{booktabs}
\usepackage{multirow}

\usepackage{listings}
\usepackage{xcolor}
\usepackage[scaled=0.85]{beramono}

\definecolor{codegreen}{rgb}{0,0.6,0}
\definecolor{codegray}{rgb}{0.5,0.5,0.5}
\definecolor{codepurple}{rgb}{0.58,0,0.82}
\definecolor{backcolour}{rgb}{0.95,0.95,0.92}
\definecolor{textblue}{rgb}{.2,.2,.7}
\definecolor{textred}{rgb}{0.54,0,0}
\definecolor{textgreen}{rgb}{0,0.43,0}
\definecolor{codered}{rgb}{201,72,12}
\definecolor{codeblue}{rgb}{0.0, 0.0, 1.0}
\definecolor{awesome}{rgb}{1.0, 0.13, 0.32}
\usepackage[T1]{fontenc}
\usepackage[scaled=0.85]{beramono} 
\definecolor{main}{HTML}{283618}    % setting main color to be used
\definecolor{sub}{HTML}{FBF8CC}     % setting sub color to be used

\definecolor{mainI}{HTML}{283618}    % setting main color to be used
\definecolor{subI}{HTML}{FFCFD2}     % setting sub color to be used

\newtcolorbox{boxK}{
    sharpish corners, % better drop shadow
    boxrule = 0pt,
    toprule = 4.5pt, % top rule weight
    enhanced,
    fuzzy shadow = {0pt}{-2pt}{-0.5pt}{0.5pt}{black!35} % {xshift}{yshift}{offset}{step}{options} 
}

\newtcolorbox{boxG}{
    enhanced,
    boxrule = 0pt,
    colback = sub,
    borderline west = {1pt}{0pt}{main}, 
    borderline west = {0.75pt}{2pt}{main}, 
    borderline east = {1pt}{0pt}{main}, 
    borderline east = {0.75pt}{2pt}{main}
}

\newtcolorbox{boxI}{
    enhanced,
    boxrule = 0pt,
    colback = subI,
    borderline west = {1pt}{0pt}{mainI}, 
    borderline west = {0.75pt}{2pt}{mainI}, 
    borderline east = {1pt}{0pt}{mainI}, 
    borderline east = {0.75pt}{2pt}{mainI}
}

\newcommand*{\Mname}{\textsc{SuperGen}}

\newcommand{\superscr}[1]{\textsuperscript{#1}}

\newcommand{\dcircle}[2][0.4]{%
  \tikz[baseline=(char.base)]{
    \node[shape=circle, draw=black, fill=black,
          minimum size=2*#1 cm, inner sep=0pt] (char)
          {\textcolor{white}{#2}};
  }%
}

\newcommand{\cmark}{\ding{51}} 
\newcommand{\xmark}{\textcolor{red}{\ding{55}}}

\begin{document}

%don't want date printed
\date{}

\title{\Large \bf \Mname{}:  An Efficient Ultra-high-resolution Video Generation System with Sketching and Tiling}

% For authors
\author{
Fanjiang Ye\superscr{1} \and
Zepeng Zhao\superscr{2} \and
Yi Mu\superscr{3} \and
Jucheng Shen\superscr{1} \and
Renjie Li\superscr{4} \and
Kaijian Wang\superscr{1} \and
Saurabh Agarwal\superscr{5} \and
Myungjin Lee\superscr{6} \and
Triston Cao\superscr{7} \and
Aditya Akella\superscr{5} \and
Arvind Krishnamurthy\superscr{8} \and
T.~S.~Eugene Ng\superscr{1} \and
Zhengzhong Tu\superscr{4} \and
Yuke Wang\superscr{1}
\\[0.8ex]   % 这里空一行，加 affiliations
\superscr{1} Rice University \quad
\superscr{2} Carnegie Mellon University \quad
\superscr{3} University of Illinois Urbana Champaign \\
\superscr{4} Texas A\&M University \quad
\superscr{5} The University of Texas at Austin \\
\superscr{6} Cisco \quad
\superscr{7} NVIDIA \quad
\superscr{8} University of Washington
\\[0.8ex] 
}

\maketitle

\begin{abstract}

Diffusion models have recently achieved remarkable success in generative tasks (e.g., image and video generation), and the demand for high-quality content (e.g., 2K/4K videos) is rapidly increasing across various domains.
However, generating ultra-high-resolution videos on existing standard-resolution (e.g., 720p) platforms remains challenging due to the excessive re-training requirements and prohibitively high computational and memory costs. 
% Existing popular T2V or I2V generation models remain capped at the upper bound of 720p resolution. 
% A popular workaround is a cascaded upsampling that first generates a 720p video and then applies video super‑resolution (e.g., SeedVR, MGLD‑VSR); unfortunately, this two‑stage process takes a long time per clip and often degrades pixel fidelity under close inspection. 
%
To this end, we introduce {\Mname{}}, an efficient tile-based framework for ultra-high-resolution video generation. 
\Mname{} features a novel training-free algorithmic innovation with tiling to successfully support a wide range of resolutions without additional training efforts while significantly reducing both memory footprint and computational complexity.
%
% \Mname{} introduces a novel tile‑based algorithmic innovation that 
%
Moreover, \Mname{} incorporates a tile-tailored, adaptive, region-aware caching strategy that accelerates video generation by exploiting redundancy across denoising steps and spatial regions.
\Mname{} also integrates cache-guided, communication-minimized tile parallelism for enhanced throughput and minimized latency. 
Evaluations show that \Mname{} maximizes performance gains while achieving high output quality across various benchmarks.
% achieves high quality scores and better runtime efficiency across various benchmarks.
% delivers a \todo{xxx}\(\times\) speedup via the caching policy, and maintains a \todo{xxx}\(\times\) scalability using tile parallelism with multiple GPUs. 
% \Mname{} is open-source anonymously at \url{https://anonymous.4open.science/r/UltraGen-4632}
\end{abstract}

\section{Introduction}
% Diffusion is widely used and popular in specific areas. But only limited to 720p, 4k is rare. Meanwhile, 4k is important in xxx areas.
Recently, diffusion models have achieved remarkable breakthroughs and widespread popularity~\cite{ho2020denoising,song2020denoising}, demonstrating unparalleled performance in generative tasks. From U‑Net architectures~\cite{blattmann2023stable} to Diffusion Transformers (DiT)~\cite{hong2022cogvideo, wan2025wan,zheng2024open, kong2024hunyuanvideo}, these transformer‑based models
%leverage scalable architecture and enhanced capacity, establishing 
have established themselves as state‑of‑the‑art paradigms across audio~\cite{wang2024av}, image~\cite{esser2024scaling}, video~\cite{wan2025wan}, gaming~\cite{{yang2024playable}}, 3D applications~\cite{mo2023dit}.
However, existing video generation models remain limited to a maximum resolution of 720p~\cite{kong2024hunyuanvideo, wan2025wan, hong2022cogvideo} and lack support for ultra-high-resolution video synthesis (e.g., 4K), which is essential for real-world applications such as film production~\cite{pecheranskyi2023attractive}, gaming~\cite{watson2020deep}, virtual reality~\cite{guo2023how}, medical analysis~\cite{umirzakova2024medical}, and scientific discovery~\cite{wu2025star}, where high visual fidelity is critical. Enabling ultra-high-resolution generation not only enhances perceptual quality and immersion but also extends generative models to professional-grade content creation.
% Moreover, performance often degrades, and the model architecture does not support out-of-scope resolution when processing resolutions that differ from those seen during training. 

% Why they can not scale to 4k: 1) Foundation model not support. Train a 4K supported model need a large amount of training datasets. 2) even though train such a model, directly do inference to generate video will still meet OOM (hardware-limitation)
Despite such increasing demand, directly generating ultra-high-resolution video remains highly challenging. A key obstacle lies in the scarcity of high-quality, real-world datasets at ultra-high resolutions for generative model training.
Similar to models trained at standard resolutions~\cite{wan2025wan,chen2025gokuflowbasedvideo}, achieving strong performance at higher resolutions would require billions of ultra-high-resolution samples (e.g., 4K videos), which is nearly infeasible at the current stage. 
Beyond the scarcity of training data, hardware limitations exacerbate the difficulty. 
Even if such a generative model capable of high-resolution video synthesis is available, inference often exceeds the memory capacity of commodity GPUs when generating a full ultra-high-resolution clip in a single pass.
Moreover, the runtime cost of producing even a short 4K video is prohibitively high. For instance, as estimated from Table~\ref{tab:video_model_perf}, generating a 128-frame 4K clip directly can take up to 40 hours on a high-end NVIDIA H100 GPU, making real-time and efficient ultra-high-resolution video generation impractical for the aforementioned applications. 
Additionally, commonly used system optimizations, such as caching mechanisms and parallelism, have not been systematically considered in the context of ultra-high-resolution video generation. This is because those existing techniques remain largely tailored to regular resolutions~\cite{wang2024pipefusion}, leaving substantial room for exploring applicability in ultra-high-resolution synthesis.

To overcome these issues, we identify several key opportunities. 
{\underline{First}}, additional training of generative models tailored for ultra-resolution generation is, in fact, unnecessary. 
Our key insight is that ultra-high-resolution video generation can be achieved in a training-free manner by leveraging pretrained models originally designed for regular resolutions.
A promising strategy is to first generate a coarse, global overview to guide the overall structure, followed by localized refinement to enrich details, akin to the process of painting.
In this setting, the pretrained model effectively acts as a set of ``brushes'' for detail refinement, provided that the global structural information is well preserved.
This enables consistent, high-fidelity video generation, even for videos with diverse aspect ratios. 
{\underline{Second}}, at the ultra-high-resolution scale, caching and parallelism techniques remain underexplored yet hold significant potential for further exploitation.
Traditional caching mechanisms exploit the high similarity of intermediate features across denoising steps to improve efficiency.
In the context of ultra-high-resolution video generation with local tile refinement, caching can be made more fine-grained and flexible.
Specifically, one can apply cache reuse not only across diffusion timesteps within each tile but also adaptively across different spatial regions, thereby further amplifying the benefits of caching.
For parallelism, beyond standard sequence and tensor parallelism ~\cite{fang2024xditinferenceenginediffusion} 
that are widely used in LLM training, the local refinement stage of the diffusion model naturally affords a form of communication-minimized tile-level parallelism, since tiles are largely independent. 
This orthogonal parallelism strategy can substantially improve scalability and efficiency for ultra-high-resolution video generation.

\begin{figure}[t] 
  % \begin{center}
\includegraphics[width=\linewidth]{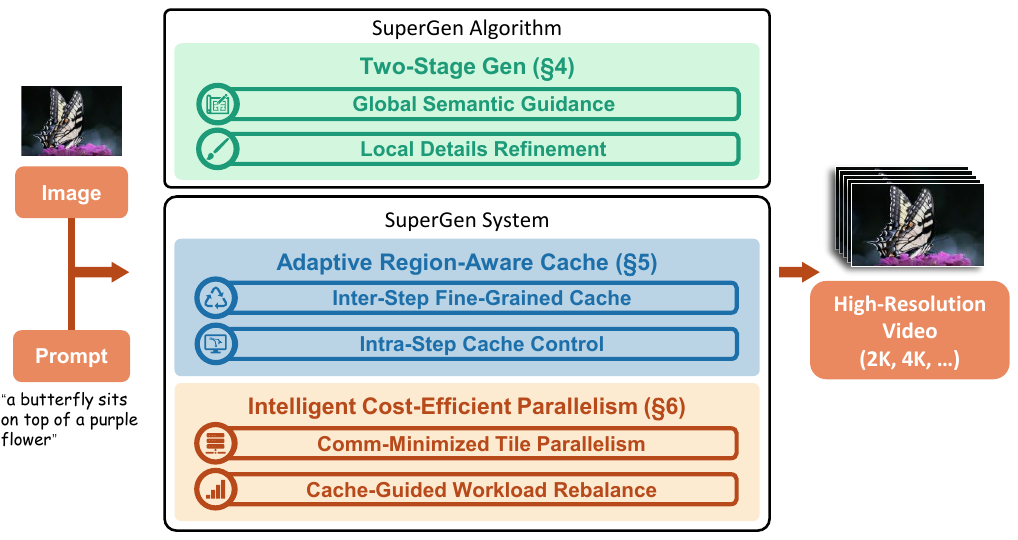}
  % \end{center}
  \vspace{-10pt}
  \caption{Overview of \Mname{}.}
  \vspace{-10pt}
  \label{fig: overview of TVG}
\end{figure}

\begin{figure}[t] \small
\begin{lstlisting}[caption={Example of a DiT-based SuperGen.}, label={lst:listing-python}]
import SuperGen
# Import other packages, e.g., PyTorch, Diffusers ...
SuperGen.config = {res="2K", unit="720p", num_GPUs=4}
# Define SuperGen tasks.
class BaseGen(SuperGen.Module):
    def __init__(self, latents, config):
        super().__init__()
        self.manager = SuperGen.setup()
        self.transformer = CachingTransformer.load(config)
        self.scheduler = FusedScheduler.setup(config)
    # Define tile-based generation function.
    def Tile_gen(self, n_steps, enable_cache, re_balance):
        for i, t in enumerate(n_steps):
            self.manager.communicate("latent")
            self.manager.shift()
            if enable_cache: # If applying cache.
                tile_pos = self.manager.get_boundary(idx)
                self.manager.skip(tile_pos, idx, thres)
        if re_balance:  # If triggering rebalance.
            self.manager.rebalance()
        # Denoise -> Fuse -> Update
        noise = self.transformer(latents, config)
        fused = self.manager.fuser(noise, config)
        denoised = self.scheduler.step(fused, latents)
\end{lstlisting} 
\vspace{-10pt}
\end{figure}

% \fanjiang{\Mname{} follows a multi-GPU design combined with caching and re-balancing, and compatible with different custom diffusion models. Listing~\ref{lst:listing-python} shows example for the tile-based video generation. There are three major components: 1) \textit{Distributed Manager} (Line 7) manages configurations, tensors, data movement, and communications, ensuring readability and maintainability, and employs \texttt{get\_boundary} to track each tile’s status and position. 2) \textit{CachingTransformer} (Line 8) exploits temporal and spatial redundancy via our caching mechanism. 3) \textit{Rebalancing} (Line 19) utilizes \texttt{allgather} to reorganize the tile-GPU configurations settings. Users will implement the \texttt{Tile\_gen} for running and easily extend it to custom diffusion models.}

Based on the above insights, we propose \Mname{}, a training-free and efficient framework for generating high-resolution videos with sketch-tile collaboration (Figure~\ref{fig: overview of TVG}). 
The core idea of \Mname{} is to emulate the workflow of an artist: the generation process follows a two-stage paradigm in which a coarse, low-resolution sketch provides a global structural overview, followed by iterative fine-grained tile-based refinement that enriches details while preserving the semantics/content of the original sketch.
% The core idea of \Mname{} is that the generation of the ultra-high-resolution video could follow a two-stage generation like an artist, where we start with the rough low-resolution sketches and outlines coarse-grained generation to guide the entire canvas overview, and followed by fine-grained tile generation via iterative detailed refinement while complying with the semantics/contents from the original sketches. 
With this key insight of sketch-tile collaboration, we develop a holistic algorithm-system co-design that exploits performance opportunities across multiple levels of the generation hierarchy, such as different regions of a video clip and different denoising steps, thereby minimizing redundant computation while maximizing the potential of parallelism for efficiency. 

% To address the fundamental bottlenecks of ultra-high-resolution diffusion, 
Specifically, \Mname{} integrates a set of complementary techniques that jointly improve efficiency, scalability, and usability. First, we adopt a training-free two-stage generation design, which decomposes full-attention denoising into parallelizable tile-level operations. This design not only avoids the cost of retraining but also dramatically increases parallelism during inference. Building on this structure, we introduce inter-step fine-grained caching to leverage temporal similarity across denoising steps, together with intra-step region-aware cache control to exploit spatial redundancy across tiles. These mechanisms substantially reduce redundant computation while preserving generation fidelity. To enable practical scaling beyond a single device, we develop a communication-minimized tile-parallel execution scheme, augmented by cache-guided workload rebalancing that mitigates stragglers and maintains low end-to-end latency. To make these capabilities accessible to developers, we further provide a unified base class, \texttt{SuperGen} (Listing~\ref{lst:listing-python}), which integrates seamlessly with the existing diffusion pipeline in the \texttt{Diffusers} framework~\cite{von-platen-etal-2022-diffusers}.
To sum up, we make the following contributions:
\begin{itemize}
    \vspace{-4pt}
    \item We propose the first training-free diffusion model design for ultra-high-resolution video generation based on two-stage resolution scaling (\S\ref{sec: algorithm framework}), which synergizes standard-resolution scene sketches with high-resolution regional details, achieving low computation and memory complexity and high outcome quality.
    % that seamlessly extends existing diffusion models to effectively support multi-resolution video generation.
    %
    \vspace{-4pt}
    \item We introduce a holistic system support tailored to our new algorithmic design, featuring fine-grained region-aware caching (\S\ref{sec:hierarchical_caching}) across diffusion steps and spatial regions of videos, and a resource-aware, cache-guided, tile-level parallelism strategy (\S\ref{sec:multi_gpu}) within each diffusion step to harvest the real performance gains.

    % to minimize the redundancy while maximizing parallelism across different levels of diffusion hierarchies. Our system 
    % adaptive region-aware caching mechanism that leverages redundancy across denoising steps and spatial semantic heterogeneity within each step, enabling fine-grained and controllable caching.

    % \item We present a resource-aware, dynamic tile-wise parallelism strategy that enables concurrent processing, substantially accelerating video generation by eliminating the inefficiencies of sequential tile-by-tile execution and offering superior scalability. 
    \vspace{-4pt}
    \item Extensive evaluation demonstrates that \Mname{} delivers state-of-the-art quality across multiple benchmarks and backends, while achieving up to $6.2\times$ speedup without compromising quality.
\end{itemize}

\section{Background}
\begin{figure}[t]
  \begin{center}
\includegraphics[width=\linewidth]{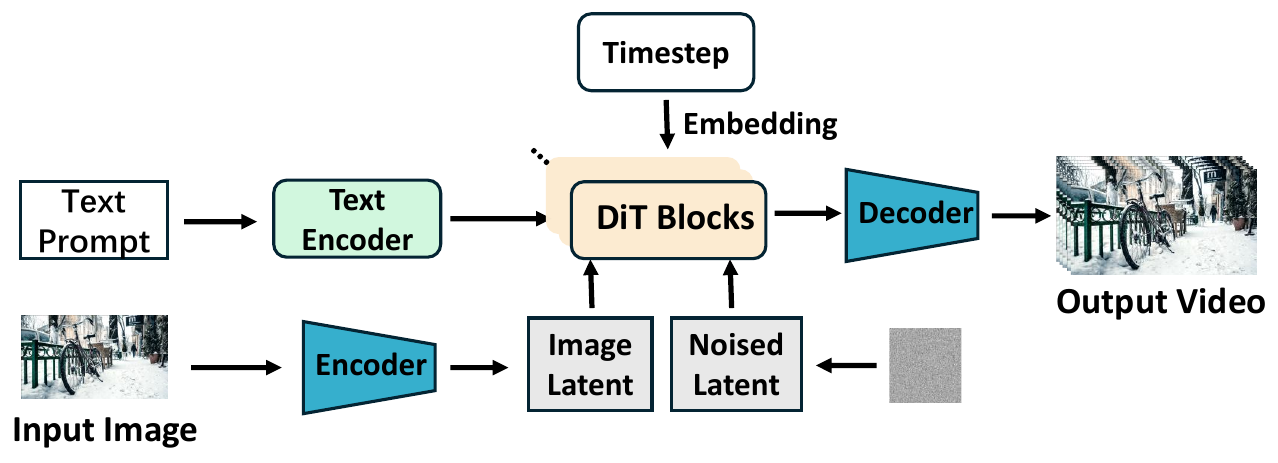}
  \end{center}
  \caption{Overview of DiT image-to-video inference process.}
  \label{fig: sec2_1_Diffusion Process}
\end{figure}
\subsection{Video Diffusion Model }
Latent Diffusion Models (LDMs) have emerged as a dominant paradigm for image synthesis and video generation, owing to their strong capability in producing high-quality outputs. These models operate in a compressed latent space, where an encoder $\mathcal{E}$ maps the input image or video $x$ to its latent representation $z = \mathcal{E}(x)$. LDMs follow a two-stage diffusion process: a forward (noising) process and a reverse (denoising) process. In the forward process, Gaussian noise is gradually added to the latent data $z_o$ over a sequence of timesteps $T$, transforming it into a sample that approximates the standard normal distribution $\mathcal{N}(0, I)$, as described in Equation~\ref{equation: forward diffusion}, where $\{\beta_t\}_{t=1\ldots T}$ is a predefined variance schedule.
\begin{equation}
\label{equation: forward diffusion}
q(\mathbf{z}_t \mid \mathbf{z}_{t-1}) = \mathcal{N} \left( \mathbf{z}_t; \sqrt{1 - \beta_t} \, \mathbf{z}_{t-1}, \beta_t \mathbf{I} \right).
\end{equation}
In the reverse process, a learnable noise prediction network is utilized to reconstruct the original data $z_o$ by iteratively denoising from the noised sample $z_T$, as outlined in Equation~\ref{equation: reverse process of diffusion}, where $\mu_\theta(\mathbf{z}_t, t)$ and $\Sigma_\theta(\mathbf{z}_t, t)$ are the predicted mean and variance of the Gaussian distribution. 
\begin{equation}
\label{equation: reverse process of diffusion}
    p_\theta(\mathbf{z}_{t-1} \mid \mathbf{z}_t) = \mathcal{N} \left( \mathbf{z}_{t-1}; \mu_\theta(\mathbf{z}_t, t), \Sigma_\theta(\mathbf{z}_t, t) \right).
\end{equation}
Among various diffusion backbones, Diffusion Transformer (DiT)~\cite{hong2022cogvideo, wan2025wan, zheng2024open, kong2024hunyuanvideo} has become particularly prominent due to its strong modeling capacity and scalability. In typical DiT-based video generation pipelines \cite{peebles2023scalable} (Figure~\ref{fig: sec2_1_Diffusion Process}), the process comprises three core components: a Variational Autoencoder (VAE), a text encoder, and a DiT model. The input image is first encoded into a latent representation by the VAE, while the text prompt is transformed into a semantic embedding via the text encoder. These inputs, including latent noise, timestamp embeddings, and textual features, are processed by the DiT model, which employs interleaved spatial-temporal or full attention mechanisms along with cross-attention to the text for semantic alignment. After iterative denoising, the refined latent is decoded by the VAE to produce the final frames.

\subsection{Ultra-high-resolution Video Generation}
\textbf{Algorithmic Support:} Ultra-high-resolution video synthesis remains highly challenging at the current stage of video generation, as existing models are typically trained only on datasets up to 720p and have little or no exposure to ultra-high-resolution content. Direct inference at unseen resolutions with such a pretrained diffusion model is generally infeasible and often leads to issues such as incomplete content or severe visual artifacts. 
A widely adopted but non-training-free approach is video restoration~\cite{wang2023mgldvsr, wang2025seedvr}, which generates high-resolution outputs by upscaling a pre-generated low-resolution video with a pretrained super-resolution model.
However, this method suffers from several drawbacks. 
First, it is not training-free and is very computationally expensive, as it requires millions of high-quality paired low- and high-resolution samples for effective training. 
Second, it is inherently constrained by fixed scaling factors: if the model has not been trained for a particular resolution, it cannot upscale to that target high-resolution, thereby limiting its flexibility in supporting diverse resolutions.

Currently, research on training-free ultra-high-resolution video synthesis remains very limited, with most prior work focusing only on images. For instance, Hidiffusion~\cite{zhang2023hidiffusion} mitigates object duplication in images by dynamically adjusting U-Net feature map sizes; Demofusion~\cite{du2024demofusion} employs skip residuals and dilated sampling to progressively upscale images. While effective in the image domain, these approaches cannot be directly extended to ultra-high-resolution video synthesis, as videos involve an additional temporal dimension. For example, the overlapping-based method applied in image synthesis~\cite{bar2023multidiffusion} exhibits pronounced boundary inconsistencies when extended to videos, primarily due to the loss of temporal coherence and the accumulation of errors around patch boundaries. This highlights the urgent need for an effective training-free algorithm framework that ensures consistency in ultra-high-resolution video synthesis.

\noindent \textbf{System Support:} {System acceleration for DiT-based video inference is also essential, as the substantial memory footprint and long runtimes caused by the full attention mechanism are often prohibitive. Existing training-free acceleration strategies primarily include caching~\cite{agarwal2024approximate,ma2024deepcache,ma2024learning,selvaraju2024fora,shenmd,chen2024delta} and parallelism~\cite{li2024distrifusion,zhang2024partially,chen2024asyncdiff,wang2024pipefusion,fang2024xditinferenceenginediffusion}. Caching leverages the high similarity of intermediate features across adjacent steps to reuse previous results and skip certain denoising steps. On the other hand, parallelism is typically not communication-efficient and is often implemented in the form of sequence or tensor parallelism, as those commonly used in LLM training. However, prior work has not investigated the opportunities specific to tile-wise ultra-high-resolution generation. 
Unlike existing approaches, tile-based generation enables fine-grained, adaptive cache control across local tiles, while tile independence naturally facilitates a cost-efficient form of tile-level parallelism orthogonal to other parallelism methods.
% Together, these properties open new opportunities for substantially improving computational efficiency.
}
\section{Motivation}
\label{sec3： motivation}
{\textbf{Exploring resolution scaling without training.} 
To synthesize videos at resolutions beyond their originally trained resolution, directly applying pretrained generative models would often lead to architectural incompatibilities~\cite{wu2025megafusion} and severe quality degradation~\cite{du2024demofusion}. Despite missing a direct ultra-high-resolution scaling solution, we could resort to indirect scaling alternatives that could also provide a promising solution.
Our key insight draws from prior training-free image synthesis methods~\cite{bar2023multidiffusion}, where we observe that pretrained models already possess sufficient generative capacity at their native resolution, which can be exploited for scaling without additional training. The critical idea is resolution composability.
By decomposing the entire ultra-resolution video canvas into standard-resolution tiles, each region can be synthesized locally and independently using the originally pretrained standard-resolution model, and then we can extend the generation to higher resolutions in a training-free manner. To achieve this, one common practice is to adopt the tile overlapping and sliding-window strategy from~\cite{bar2023multidiffusion} to video synthesis. 
Unfortunately, such practices lead to pronounced boundary inconsistencies and severe semantic incompleteness, since the lack of information exchange across tiles results in insufficient or incoherent global semantics. 
% \textcolor{red}{add a example of video frames or score degradation for illustration}. 
%
This limitation drives the demand for a solution that can effectively unify global structural consistency with local detail refinement.
\begin{table}[t] \small
\footnotesize
\centering
\caption{Comparison of maximum supported resolution, GPU memory requirements, and inference time for generating videos on an H100 GPU.}
% \vspace{-10pt}
\resizebox{\linewidth}{!}{
\begin{tabular}{lcccr}
\toprule
Model & Max Resolution & Frames & VRAM & Latency \\
\midrule
CogVideoX1.5 5B~\cite{hong2022cogvideo} & $1360\times768$ & 80 & 40GB & 400s  \\
HunyuanVideo 13B~\cite{kong2024hunyuanvideo} & $1280\times720$ & 128 & 70GB & 1,800s 
\\
% Wan2.1 14B~\cite{wan2025wan} & $1280\times720$ & 128 & 80GB & 3000s \\
\bottomrule
\end{tabular}}
\label{tab:video_model_perf}
\end{table}

\vspace{2pt}
\noindent {\textbf{Exploiting system opportunities with tiling}. While the above two-stage framework enables scaling to ultra-high resolutions, it also magnifies efficiency as a new bottleneck. State-of-the-art diffusion-based video generators~\cite{kong2024hunyuanvideo, wan2025wan} require tens of minutes to synthesize a 128-frame video at a standard resolution of 1280$\times$720, and scaling to 4K resolution (3840$\times$2160) extends the runtime to several hours or even tens of hours. For example, extrapolations from Table~\ref{tab:video_model_perf}: 
\[
T_{4K} \approx 1800\text{s} \times (\tfrac{3840 \times 2160}{1280 \times 720})^2 \,/\, 3600 \approx 40 \text{ h},
\]
show that generating even a single 128-frame 4K video with the state-of-the-art HunyuanVideo~\cite{kong2024hunyuanvideo} framework would take more than a day on a high-end NVIDIA H100 GPU. Such prohibitively high time consumption underscores the necessity of tailored acceleration strategies. Among possible directions (e.g., distillation and pruning) for accelerating diffusion, cache-based and parallel acceleration have gained most attention~\cite{liu2024timestep, li2024distrifusion,fang2024xditinferenceenginediffusion}, since they are training-free, easy to integrate, and effectively preserve generation quality. Yet in the context of tile-wise ultra-high-resolution video generation, these two strategies remain largely underexplored. 
Compared with the standard resolution, ultra-high resolution imposes the following challenges:
% The tile-wise setting enables finer-grained control in two aspects: 
1) when to apply caching or parallelism along the denoising trajectory, in order to minimize redundant computation across different stages, 
and 2) how to apply them at different granularities (e.g., inter-step or intra-step), which helps maximize parallelism while preserving output quality.

\begin{figure}[t]
  \begin{center}
  \includegraphics[width=0.9\linewidth]
  {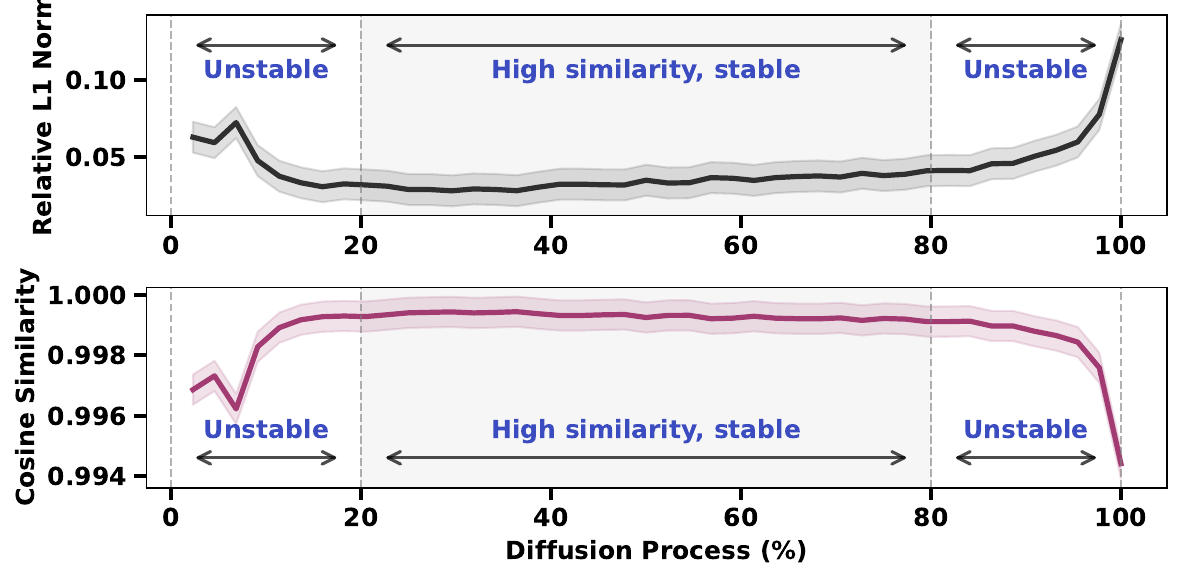}

  \end{center}
  \caption{Similarity of output predicted noise $O_t$ across adjacent denoising steps. \textbf{Top}: relative L1 distance and \textbf{Bottom}: cosine similarity. The definitions can be found in Equation~\ref{equa: similarity of predicted noise}. Predicted noise reveals strong consistency and high similarity in both magnitude and direction during intermediate steps, while unstable at the beginning and the end. }
  % of the process~\cite{liu2024timestep, yu2025ab}. }
  \label{fig:noise_prediction_similarity}
\end{figure}

\vspace{2pt}
% \noindent 
\noindent {\textbf{Minimizing redundancy with hierarchical caching}. To concretize the above efficiency opportunities, we first examine where redundancy arises in diffusion inference. 
% Prior studies~\cite{wimbauer2024cache, liu2024timestep} highlight that consecutive denoising steps often yield highly similar noise predictions. 
As shown in Figure~\ref{fig:noise_prediction_similarity}, we observe that: the predicted noises $\mathbf{O}_t$ exhibit strong similarity in both magnitude and direction (measured by relative L1 norm and cosine similarity defined in Equation~\eqref{equa: similarity of predicted noise}), indicating that intermediate results contain substantial reuse potential across timesteps.
\begin{equation} \small
\label{equa: similarity of predicted noise}
\mathrm{L1}_{\mathrm{rel}}(\mathbf{O}, t) = \frac{\left\| \mathbf{O}_t - \mathbf{O}_{t+1} \right\|_1}{\left\| \mathbf{O}_{t+1} \right\|_1}, \mathrm{CosSim}(\mathbf{O}, t) = \frac{\langle \mathbf{O}_t, \mathbf{O}_{t+1} \rangle}{\left\| \mathbf{O}_t \right\|_2 \left\| \mathbf{O}_{t+1} \right\|_2}
\end{equation}
Moreover, recent findings~\cite{liu2025regionadaptivesamplingdiffusiontransformers} indicate that diffusion models devote more updates to semantically salient regions during sampling, whereas large background regions evolve more slowly. This insight suggests a spatial opportunity: caching need not be applied uniformly across the canvas, but can be adjusted depending on regional dynamics across tiles.
These observations motivate a hierarchical caching strategy that minimizes both inter-step and intra-step redundancy, improving efficiency without sacrificing quality.}

% \noindent
\noindent \textbf{Maximizing parallelism with tile-level independence}. While caching can substantially reduce redundant computation on a single GPU, such optimizations alone are insufficient for scaling to ultra-high resolutions. Our profiling with NVIDIA Nsight Compute shows that key diffusion operators (e.g., FlashAttention~\cite{dao2022flashattention}) already saturate the available SMs on a single GPU, implying that further acceleration requires scaling across multiple GPUs.
Within the tile-wise generation setting, the local refinement process naturally lends itself to parallelization: attention incurs no Query–Key–Value communication across tiles, exposing opportunities for lightweight tile-level parallelism that reduces attention complexity.
Besides, static workload allocation may lead to imbalance (e.g., when distributing nine tiles across eight GPUs), and cache-induced variability can leave some devices idle while others are active. This observation suggests that effective parallelism at ultra-high resolution must not only exploit tile independence but also incorporate cache-driven workload balancing to fully realize scalable tile-level parallelism.

% Section 4

\section{Training-free Two-stage Generation}
\label{sec: algorithm framework}

\begin{figure}[t]
  \begin{center}
\includegraphics[width=1.0\linewidth]{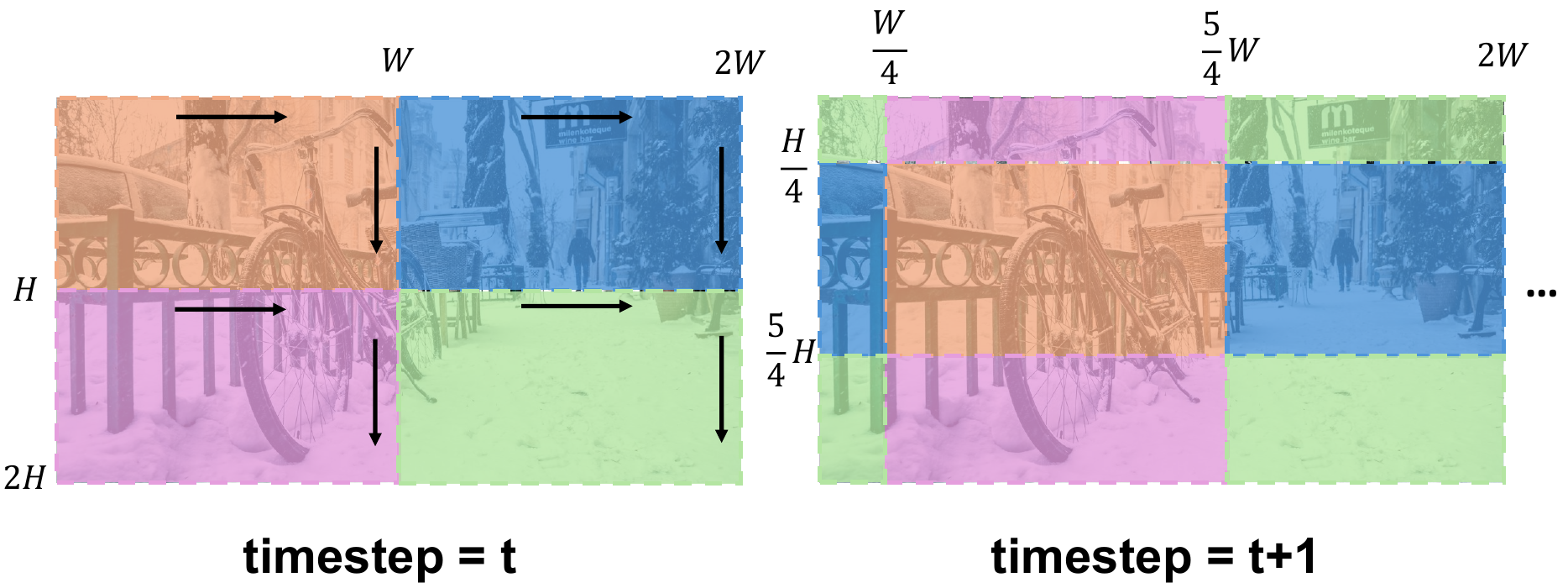}
  \end{center}
\vspace{-10pt}
  \caption{Illustration of tile shifting (2K canvas is covered with 4 tiles of 720p).
  }
  \label{fig: sec_4_2_2_tile_shifting mechanism}
  \vspace{-5pt}
\end{figure}

\begin{algorithm}[t] \small
\caption{Two-Stage High-Resolution Video Generation}
\label{alg:Tile-Based Resolution Scaling Framework}
\begin{algorithmic}[1]
\Require Input image $x$, text prompt $p$, denoising steps $T$, tile size $s$, renoise steps $k$.
\Ensure High-resolution generated video $\hat{v}$.
% \vspace{0.5em}
\Statex \textcolor{blue}{--- \textbf{Stage 1}: Low-resolution generation ---}
\State $\ell \gets \textsc{Model}(x, p)$ 
% \vspace{0.5em}
\Statex \textcolor{blue}{--- Upscaling to high-resolution latent ---}
\State $v \gets \textsc{Model.Decode}(\ell)$ \Comment{Convert latent to pixel space}
\State $\tilde{v} \gets \textsc{Interpolate}(v)$ \Comment{Upscale in pixel space}
\State $\tilde{\ell} \gets \textsc{Model.Encode}(\tilde{v})$ \Comment{Convert back to latent space}
\State $L \gets \textsc{AddNoise}(\tilde{\ell}, T, k)$ \Comment{Re-noise the video latent}
% \vspace{0.5em} 
\Statex \textcolor{blue}{--- \textbf{Stage 2}: High-resolution refinement with tiling ---}
\State $\mathcal{P} \gets \textsc{PartitionTiles}(\text{shape}(L), s)$
\For{$t = T-k$ to $T$}
    \If{\textsc{ShiftThisStep}$(t)$}
        \State $\mathcal{P} \gets \textsc{Shift}(\mathcal{P})$ \Comment{Shift tiles to new positions}
    \EndIf
    \State $\mathcal{N} \gets [\,]$ \Comment{List of noise tiles}
    \ForAll{$\tau \in \mathcal{P}$}
        \State $L_\tau \gets \textsc{SliceTile}(L, \tau)$
        \State $n_\tau \gets \textsc{Model.DiT}(L_\tau, p, t)$ \Comment{Predict noise}
        \State Append $n_\tau$ to $\mathcal{N}$
    \EndFor
    \State $n \gets \textsc{Fuse}(\mathcal{N})$ \Comment{Restore tiles to complete noise}
    \State $L \gets \textsc{Model.Denoise}(L, n)$
\EndFor
% \vspace{0.5em}
\Statex \textcolor{blue}{--- Final decoding ---}
\State $\hat{v} \gets \textsc{Model.Decode}(L)$
\State \Return $\hat{v}$
\end{algorithmic}
\end{algorithm} 

% Section 4.1
% TVG framework Figure
\begin{figure*}[t]
  \centering
    \includegraphics[width=\linewidth, keepaspectratio]{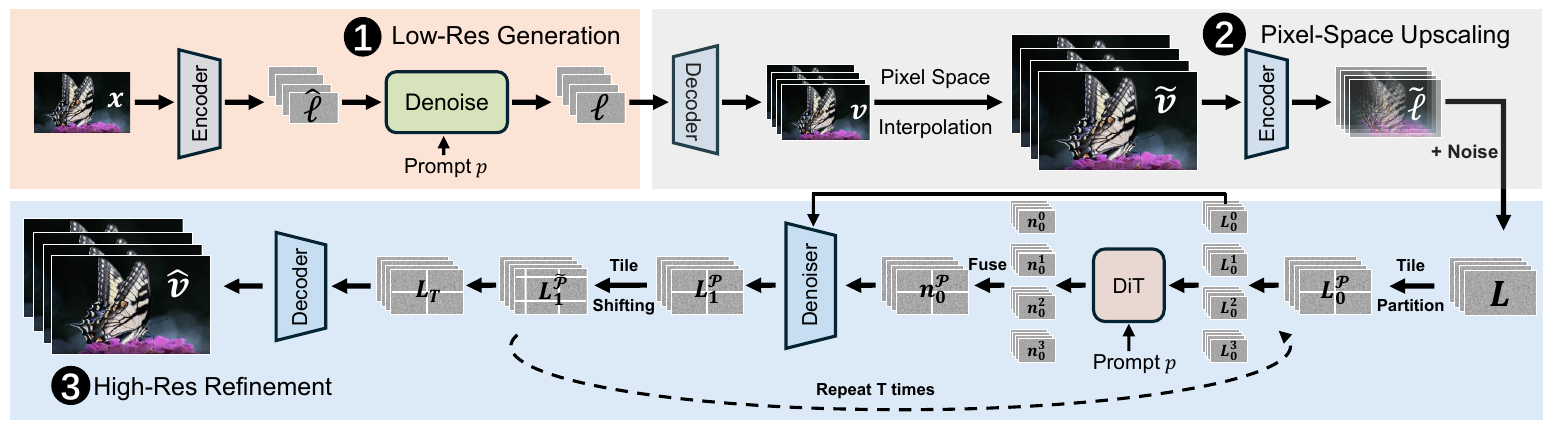}
  \vspace{-20pt}
  \caption{Illustration of tile-based resolution scaling framework.}
  \vspace{-10pt}
  \label{fig: sec4_2_TVG_framework}
\end{figure*}

We propose our training-free two-stage generation 
% \mlee{while reading eval section, I got confused with stage and phase. stage 1 has the first two phases and stage 2 has the third phase. This seems a bit confusing. maybe changing phase to step help lessen the confusion? I don't know}
framework, illustrated in Figure~\ref{fig: sec4_2_TVG_framework} and detailed in Algorithm~\ref{alg:Tile-Based Resolution Scaling Framework}. In phase 1, the original pipeline is used to generate a low-resolution latent $\ell$ from the input image $x$ and prompt $p$. In phase 2, the latent is decoded into pixel space and upscaled to the target resolution by interpolation. The upscaled result is then re-encoded into the latent space and perturbed with noise up to timestep $T-k$. In phase 3, the latents are partitioned into predefined tile sizes. Within the main loop, we perform $k$ iterations 
% \mlee{$k$ is a hyperparameter. if we do a sensitivity study on this, better to leave a reference to the evaluation section?} 
of tile-aware denoising. {We choose $k$ a bit smaller than $T$ to balance the quality and efficiency (\S\ref{sec: implementation}}). At each iteration, noise is predicted tile by tile and then fused to form the complete noise estimate. 
% Importantly, denoising is carried out using the fused holistic noise and latent representation rather than individual tiled components. 
To ensure consistency across tile boundaries, tile positions are shifted according to user-specified settings. Finally, the denoised latent is decoded to produce the high-resolution video.

%%%%%%%%%%%%%%%%%
% Section 4.2
\subsection{Global Semantic Guidance}
\label{subsec: Global Semantic Guidance}
Zero-shot generation (w/o phase \dcircle[0.18]{\textbf{1}}) of high-resolution videos via a tile-wise sampling process often introduces artifacts such as duplicated objects, primarily because individual tiles lack global semantic context, examples illustrated in Appendix~\ref{appendix: Visualization of Claims}-Figure~\ref{fig: sec_4_2_1}(c). To mitigate this issue, rather than directly synthesizing a high-resolution video, we first generate a low-resolution video to serve as a global reference (Line 1 in Algorithm~\ref{alg:Tile-Based Resolution Scaling Framework}). 
% Before entering the phase \dcircle[0.18]{\textbf{3}}, 
The video is then upscaled to produce a clean high-resolution reference latent in phase \dcircle[0.18]{\textbf{2}}, which is then re-noised to preserve the original structural information (Line 2-5). Empirically, we observe that structural coherence improves substantially when the latent representation is initialized with a deterministic amount of noise
% \mlee{how do we pick noise deterministically? what type of noise is it if not Gaussian? some details seems to be missing.}
rather than purely random Gaussian noise before entering phase \dcircle[0.18]{\textbf{3}}. Random initialization often yields outputs with noticeable inconsistencies, whereas controlled noise injection retains critical structural information from the reference, enabling more accurate detail refinement.
Furthermore, we find that performing interpolation directly in the latent space—rather than in the pixel space~\cite{liu2024dynamicscaler, du2024demofusion} introduces visible defects (detailed in Appendix~\ref{appendix: Visualization of Claims}-Figure~\ref{fig: sec_4_2_1}(a) and (b)). This occurs because encoding a scaled or rotated image does not correspond to scaling or rotating its original latent representation~\cite{kouzelis2025eq}.
%%%%%%%%%%%%%%%%%%%%%%%
% Section 4.3
\subsection{Local Details Refinement}
In the refinement stage (Line 6-19), the video canvas is partitioned into multiple non-overlapping tiles by slicing along the spatial dimensions (height and width), while preserving the full temporal sequence of frames within each tile. Each tile is then processed independently using the regular-resolution model to predict its local noise. Prior to the scheduler update, the predicted noise from all tiles is aggregated to ensure that the entire canvas adheres to a consistent denoising trajectory. This step is crucial because higher-order samplers require coherent historical states, which would otherwise be disrupted by tile shifting. After the scheduler update, the denoised tiles are reassembled, and tile shifting is applied before proceeding to the next denoising step. 
% \subsection{Consistency Preserving via Shifting}

Previous tile-based generation approaches, such as MultiDiffusion~\cite{bar2023multidiffusion}, maintain fixed tile positions throughout the generation process. While this strategy proves effective for image synthesis, it fails to maintain temporal consistency in video synthesis. Motivated by~\cite{frolov2025spotdiffusion}, we introduce a novel tile shifting mechanism (Figure~\ref{fig: sec_4_2_2_tile_shifting mechanism}). The core insight underlying this design is that boundary artifacts introduced at one timestep can be corrected in subsequent steps through strategic tile sliding. Unlike~\cite{frolov2025spotdiffusion}, which applies random tile shifts, our method adopts deterministic shifting along both horizontal and vertical directions with a fixed stride. This design enables the use of non-overlapping tiles while still ensuring coherent transitions across the spatial domain, thereby significantly reducing the computational cost of the denoising process.
% \begin{figure}[t]
%   \begin{center}
% \includegraphics[width=1.0\linewidth]{figures/sec_4_2_2_effect_of_shifting-v3.pdf}
%   \end{center}
%   \caption{\fanjiang{Remove figure 7}Illustration of the effect of tile shifting. \textbf{Left}: Without tile shifting, noticeable inconsistencies appear along tile boundaries. \textbf{Right}: With tile shifting applied, the generated video exhibits seamless transitions and improved spatial consistency across tiles.
%   }
%   \label{fig: sec_4_2_2_tile_shifting effect}
% \end{figure}
%%%%%%%%%%%%%%%%%%%%%%%%%%%%%%%%%%%%%%%%%%%%%%%%%%%%%%%%%%%%%%%%%%%%%%%%%%%%%%%%%%%%%%%%%%%%%%
% Section 5

\section{Fine-grained Region-aware Cache} 
\label{sec:hierarchical_caching}

% \mlee{I think this section is about an optimization technique in phase 3 of Fig 5, specifically the "looping k times" part. However, the notation ($L_t$) used in Fig 5 is different from what is used ($I_t$) here. a bit confusing.} \fanjiang{Fixed. Here Latent L is corresponding to input I, the predicted noise n is correspoding to output O. I will explain the notation in section 5 to make it more clear}

% \mlee{also it may be nice to reuse some part of Fig 5 (e.g., DiT or denoiser) in Fig 6 to show how the things in fig 6 relate to the overall framework (i.e., fig 5).} \fanjiang{Fixed. Add a sentence to point out Figure 6 mainly accelerate the phase 3 in figure 5 and reuse the DiT engine and Denoiser of figure 5 in Figure 6.}

%%%%%%%%%%%%%%%%%%%%%
% Section 5.1
This section will present the architecture of our fine-grained region-aware caching system (Figure~\ref{fig:cache_system_overview}), accelerating the generation of Phase 3 in Figure~\ref{fig: sec4_2_TVG_framework}.
% , which dominates about 80\% runtime in Figure~\ref{fig: breakdown_stages}
The system pipeline manages the trade-off between computational efficiency and visual fidelity: the Input $I_t$ 
% {(which is the $L_t$ in Figure~\ref{fig: sec4_2_TVG_framework})}
is decomposed into distinct tiles before applying the region dynamicity analysis. After each tile obtains its adaptive threshold $\tau_i$, it can decide whether to reuse the cache residual or recompute via the Error Calculator. The Tile Fuser will aggregate all output $O_{t,i}$
% {(which is $n^{i}_{t}$ in Figure~\ref{fig: sec4_2_TVG_framework})}
from all parallel devices together and updates the input latent $I_{t+1}$ for next iteration.

% these key components : 1) \textit{Tile Dispatcher}, which decomposes the input latent $I_{t}$ {(which is the $L_t$ in Figure~\ref{fig: sec4_2_TVG_framework})} into spatially distinct tiles $I_{t,t}$, enabling fine-grained parallelism and localized decision-making. 2) \textit{Region Dynamicity Analyzer}, which analyzes the local content dynamics to assign an adaptive error threshold $\tau_i$, ensuring that static background regions can aggressively skip computations while dynamic foregrounds retain high precision (detailed in \S\ref{subsec: Intra-step Region-aware Cache Control}). 3) \textit{Error Calculator}, which interacts with the \textit{Cache Store} to estimate the accumulated error $E_t$ since the last update. Then the \textit{Threshold Comparator} determines whether to reuse the cached residual or trigger a full recomputation via the normal {\textit{DiT Engine}}. 
% \mlee{is denosing engine the same as denoiser in fig 5? it's not easy to connect phase 3 of fig 5 to fig 6.} \fanjiang{Here denoising engine is DiT in figure 5, fixed.} 
% 4) \textit{Tile Fuser} aggregates the output predicted noise $O_{t,i}$ {(which is $n^{i}_{t}$ in Figure~\ref{fig: sec4_2_TVG_framework})} from all parallel devices together and updates the input latent $I_{t+1}$ for next iteration.

\subsection{Inter-step Tile-wise Cache Policy}
Our design is inspired by the inherent redundancy across iterative denoising steps.
Figure~\ref{fig:noise_prediction_similarity} shows that the predicted noise exhibits strong similarity across intermediate timesteps.
This suggests that the predicted noise $O_t$ at the timestep t can be reused from the previous step, thereby skipping redundant computation. However, directly substituting $O_{t+1}$ with $O_t$ inevitably introduces substantial errors, which may accumulate over multiple steps and degrade video quality. 
% To more effectively exploit inter-step redundancy, we construct a more accurate approximation that captures how the predicted noise evolves with respect to the latent input.
\begin{figure}[t]
  % \begin{center}
  \includegraphics[width=\linewidth]{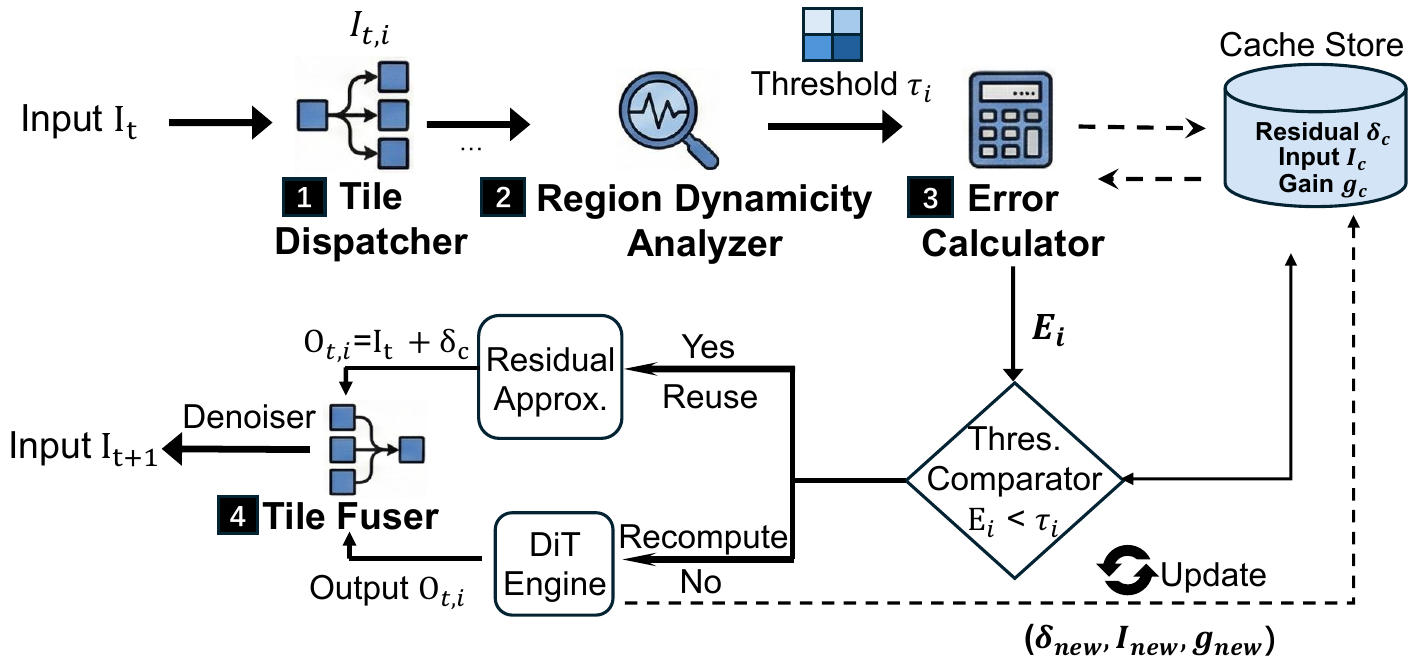}
  % \end{center}
  \vspace{-10pt}
  \caption{Fine-grained region-aware cache framework. 
  % First, \textit{Tile Dispatcher} decomposes the input latent $I_{t}$ into spatially distinct tiles $I_{t,t}$. Then, \textit{Region Dynamicity Analyzer} analyzes the local content dynamics to assign an adaptive error threshold $\tau_i$. After \textit{Error Calculator} interacts with the \textit{Cache Store} to estimate the accumulated error $E_t$ since the last update, the \textit{Threshold Comparator} determines whether to reuse the cached residual or trigger a full recomputation via the normal {\textit{DiT Engine}}. Finally, \textit{Tile Fuser} aggregates different tiles before the update in the Denoiser.
  }
  \label{fig:cache_system_overview}
  \vspace{-10pt}
\end{figure}

To better leverage this redundancy, we propose an improved approximation method that exploits the correlation between the predicted noise and the input.
% To better exploit this redundancy, we approximate the predicted noise relative to the latent input.
Instead of directly reusing $O_{t}$, we introduce cache residual $\delta_t \triangleq O_t-I_t$, which quantifies the deviation between the predicted noise and input latent. As shown in Figure~\ref{fig:similarity of residual}, we profile $\delta_t$ using the relative L1 distance, $\mathrm{L1}_{\mathrm{rel}}(\mathbf{\delta}, t)$, and cosine similarity, $\mathrm{CosSim}(\mathbf{\delta}, t)$. The results reveal that $\delta_t$ maintains stable magnitude and direction across intermediate timesteps, suggesting that we can cache a representative residual $\delta_c$ (where $c$ denotes the recomputation timestep) in the \textit{Tile-wise Cache Store}. Consequently, the predicted noise at subsequent timesteps can be approximated as $O_t\approx I_t+\delta_c$ via \textit{Residual Approximation}.

To avoid overly reuse, we track the cumulative error of predictions since the last recomputation. Specifically, starting from the last recomputed step $c$, we define the accumulated error up to timestep $t$ as:
% \mlee{minor issue: we used $k$ as \# of iterations in phase 3. here it's used as a variable.}\fanjiang{Fixed}
\begin{equation} \small
E_{c \to t} \;=\; \sum_{z=c+1}^{t} \big\| O_z - O_{z-1} \big\| .
\label{equation: accumulated error}
\end{equation}
\begin{figure}[t]
  \begin{center}
  \includegraphics[width=0.9\linewidth]{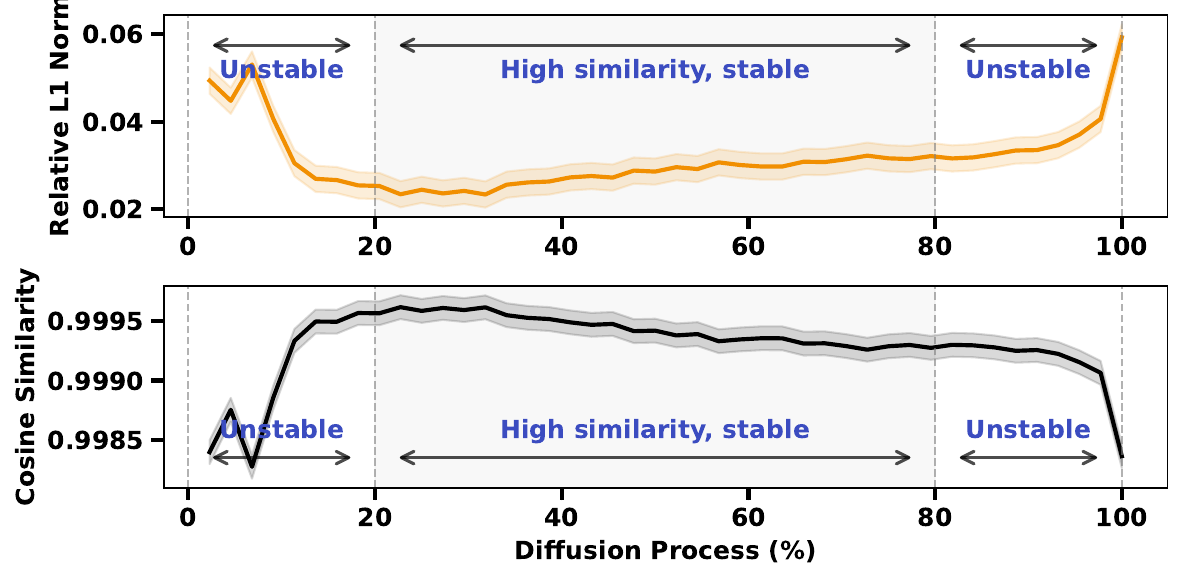}
  \end{center}
  \vspace{-10pt}
  \caption{
  The similarity of cache residue $\delta_t=O_t-I_t$ across adjacent denoising steps. \textbf{Top}: relative L1 distance and \textbf{Bottom}: cosine similarity, revealing strong consistency in both magnitude and direction across intermediate timesteps.
  % \vspace{-0.5cm}
  }
  \label{fig:similarity of residual}
\end{figure}
% Eq.~\eqref{equation: accumulated error}.
When this accumulated error $E_{c \to t}$ exceeds a predefined threshold $\tau$, the cache is refreshed by recomputing $O_t$. This adaptive strategy balances efficiency and accuracy by skipping redundant computations while preventing excessive error accumulation. However, the error calculation in Equation~\eqref{equation: accumulated error} still contains the unknown $O_t$. 
To solve this, we introduce the \textit{Gain} $g_t$, which approximates the change in the output with respect to the input, defined as
% \mlee{$\mathbf{O}_t$ and $O_t$ are confusing. Also, Fig 6 doesn't show $O_t$. I guess $O_t$ is calculated in denoising engine and only used to update the cache. maybe better to augment $O_t$ in fig 6 somehow to help understanding?}\fanjiang{$\mathbf{O}_t$ and $O_t$ are the same thing. I fixed the $\mathbf{O}_t$. $O_t$ is the output after the DiT Engine, already shown in Figure 6.}
\begin{equation} \small
g_t = \frac{\left\| O_t - O_{t-1} \right\|}{\left\| I_t - I_{t-1} \right\|}.
\label{equation:transformation_rate}
\end{equation}
From our profiling in Figure~\ref{fig:k_value_similarity}, we observe that $g_t$ remains relatively stable during intermediate denoising steps. This stability enables \textit{Error Calculator} 
% \dsquare[0.17]{\textbf{3}} 
to approximate the accumulated error $E_{c\to t}$ without explicitly computing $O_t$, as
\begin{equation}\small
E_{c\to t}
\;\approx\;
\sum_{z=c+1}^{t} g_c\,\big\| I_z - I_{z-1} \big\|
\;=\;
g_c \sum_{z=c+1}^{t} \big\| I_z - I_{z-1} \big\|
\;=\;
g_c \, L_{c\to t},
\label{eq:approx_E_using_kc}
\end{equation}
where \(L_{c\to t} \triangleq \sum_{z=c+1}^{t} \| I_z - I_{z-1} \|\) denotes the latent path length from step \(c\) to step \(t\).
Based on the approximation in Equation~\eqref{eq:approx_E_using_kc}, \textit{Threshold Comparator} determines whether to reuse the residual cache or recompute $O_t$ via \textit{Denosing Engine} according to the following rule:
\begin{equation}\small \begin{aligned} &\text{if } \; g_c\,L_{c\to t} < \tau &&\Rightarrow \;\; \text{reuse cache at step } t, \\[4pt] &\text{else } \; g_c\,L_{c\to t} \ge \tau &&\Rightarrow \;\; \text{recompute } O_t \text{ and set } c \leftarrow t . \end{aligned} \label{eq:recompute_rule_if_else} \end{equation}
Moreover, we extend the inter-step cache strategy to operate on each individual tile with \textit{Tile Dispatcher} 
% \dsquare[0.17]{\textbf{1}} 
and \textit{Tile Fuser} 
% \dsquare[0.17]{\textbf{4}} 
within our framework. 
Unlike prior work~\cite{liu2024timestep}, which makes caching decisions at the level of the entire canvas, 
our approach enables finer-grained control by performing cache at the tile level. Besides, our cache avoids intensive offline profiling to obtain the coefficient like~\cite{liu2024timestep}, which is much more compatible and applicable to other models (like U-Net based model). This localized strategy not only improves computational efficiency but also better preserves visual quality.

\begin{figure}[t]
  \begin{center}
  \includegraphics[width=0.9\linewidth]{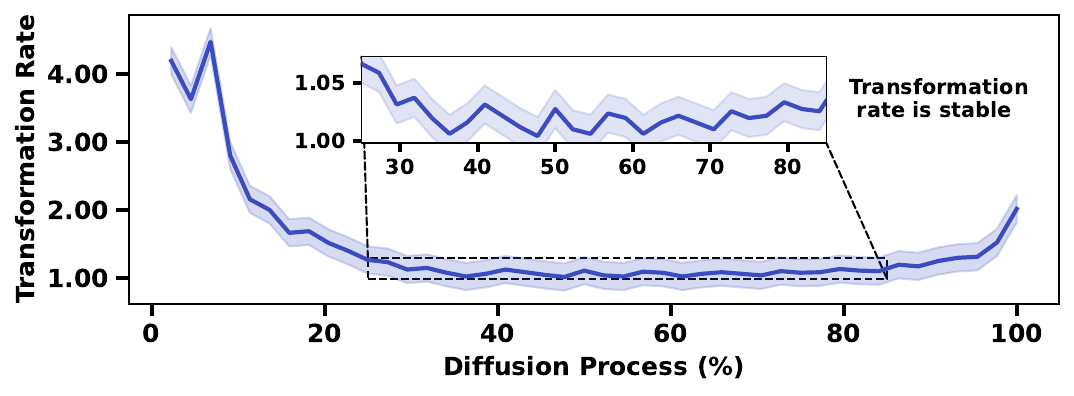}
  \end{center}
  \vspace{-10pt}
  \caption{Similarity of gain $g_t$, revealing strong consistency across intermediate timesteps.
  % \vspace{-0.5cm}
  }
  \label{fig:k_value_similarity}
\end{figure}

\subsection{Intra-step Region-aware Cache Control}
\label{subsec: Intra-step Region-aware Cache Control}
Beyond inter-step caching across denoising steps, we further exploit intra-step 
% \mlee{optimization(?)}
{optimizations} through adaptive region-aware control using \textit{Region Dynamicity Analyzer}.
% \dsquare[0.17]{2}
Rather than applying a uniform threshold, each tile is assigned its own threshold $\tau_i$ based on regional dynamics. Static background regions can tolerate larger thresholds and thus skip more steps, whereas dynamic foregrounds require smaller thresholds to avoid excessive skipping~\cite{fan2024adadiffsradaptiveregionawaredynamic}.
This design adaptively adjusts computation across tiles, improving efficiency while maintaining generation quality {(Figure~\ref{fig: cache_threshold_effect})}. 
% \mlee{if any sensitivity study on fixed threshold vs adaptive threshold is done, better to leave a forward reference?}\fanjiang{Fixed.}

To accurately distinguish static from dynamic regions, we find that the standard deviation of the predicted noise $\mathit{std}(O_t)$ serves as an effective indicator. In our visualization experiment, we found that fast-update regions typically exhibit lower standard deviation (Figure~\ref{fig:standard-deviation-of-predicted-noise}), likely because the fast-update region tends to retain more deterministic information compared to the slow-update region in each step~\cite{liu2025regionadaptivesamplingdiffusiontransformers}. After identifying the dynamicity of each tile, we adjust cache thresholds according to the corresponding standard deviation values, outlined in Appendix~\ref{appendix:algo-2}.

\begin{figure}[t]
  \begin{center}
  \includegraphics[width=\linewidth]{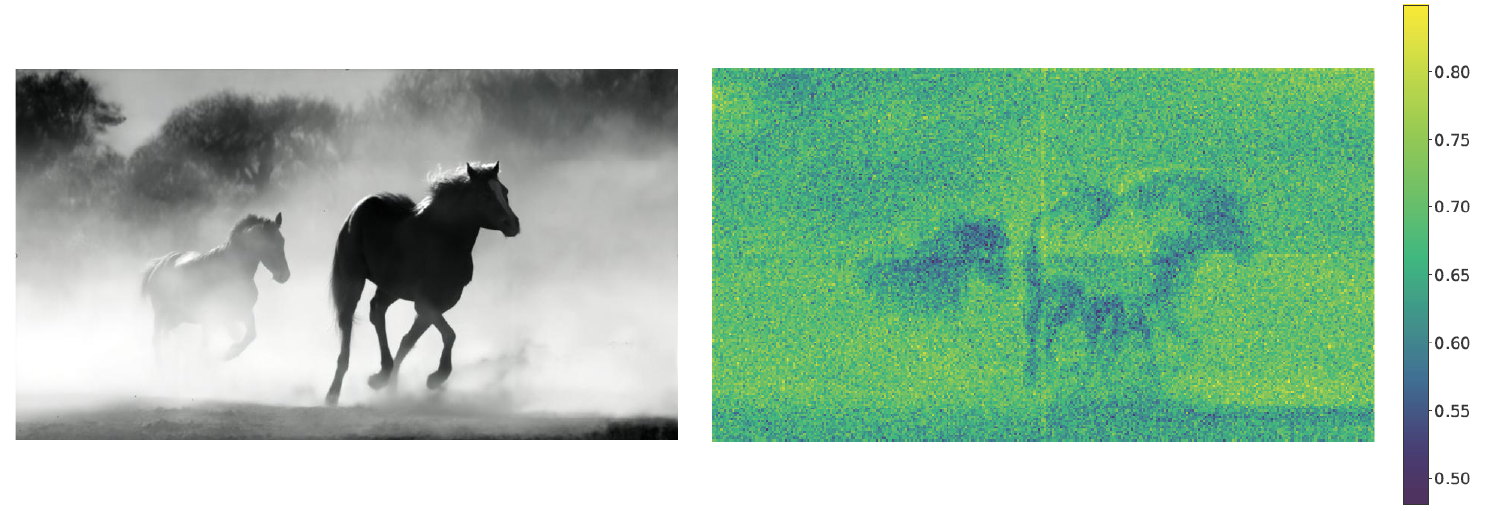}
  \end{center}
\vspace{-10pt}
  \caption{Visualization of standard deviation of noise in the 20th sampling steps (out of 50) of CogvideoX. % \vspace{-0.3cm}
  }
  \label{fig:standard-deviation-of-predicted-noise}
\end{figure}

%%%%%%%%%%%%%%%%%%%%%%%%%%%%%%%%%%%%%%%%%%%%%%%%%%%%%%%%%%%%%%%%%%%%%%%%%%%%%%%%%%%%%%%%%
% Section 6
\section{Intelligent Cost-efficient Parallelism} 
\label{sec:multi_gpu}

We propose our cache-guided, communication-minimized tile parallelism framework in Figure~\ref{fig:sec4_4_tile_parallellism}. At timestep T, Tile Dispatcher evenly divides latent across 4 GPUs. After Cache Predictor predicts which tiles can reuse cache residual, the Tile Rebalancer will automatically balance the workload (e.g., transfer one tile from GPU-3 to GPU-0), reducing the bottleneck. After each GPU goes through DiT and obtains the corresponding predicted noise, we use \texttt{allgather} to stitch all tiles and update the latent for the next timestep T+1 in Denoise Scheduler.
% Only take the single request as an example to illustrate the tile parallelism.

\begin{figure*}[t]
  \begin{center}
\includegraphics[width=1.0\linewidth]{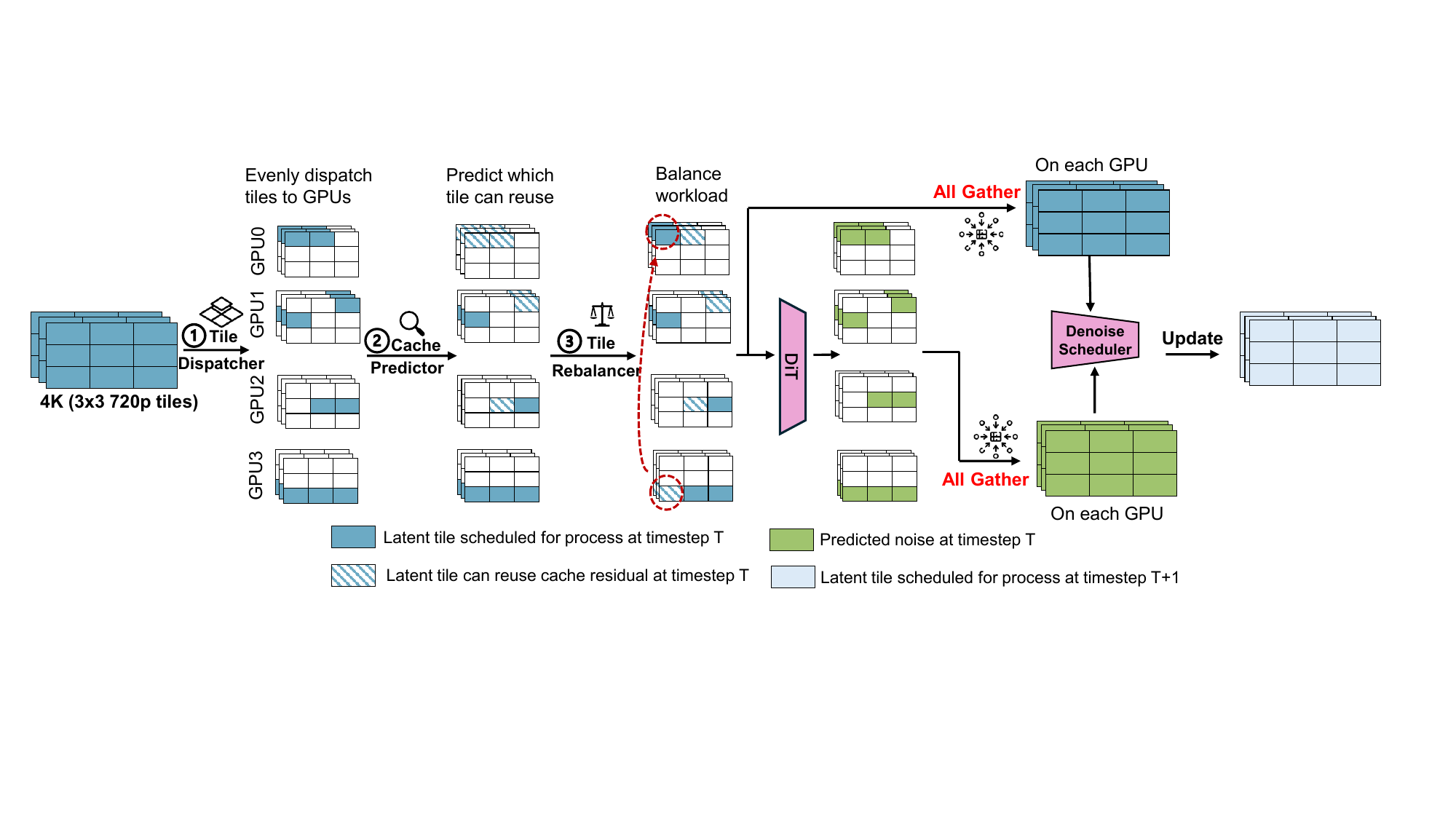}
  \end{center}
  \vspace{-10pt}
  \caption{Illustration of tile parallelism on 4 GPUs with workload rebalance and cache enabled, depicting the denoising process of 4K videos (9 tiles) from timestep T to timestep T+1. In this demonstration (similar pattern also applied to other GPU pairs), the most efficient rebalancing is to transfer one tile's workload from GPU-3 to GPU-0, as shown by the dotted red line and circle. }
  \label{fig:sec4_4_tile_parallellism}
\end{figure*}

\subsection{Comm-minimized Tile Parallelism}
\label{sec_6-1}
In our framework, tiles can be processed in DiT independently at every timestep. It allows tiles to be distributed across multiple devices, accelerating the computation-intensive denoising process. Compared with other parallelism methods, such as xDiT~\cite{fang2024xditinferenceenginediffusion}, which splits the tensors across all devices and requires all ranks to cooperatively handle a global denoising step via intensive communication, our tile parallelism focuses exclusively on local tiles and avoids heavy communication within DiT modules.
% Because it requires only minor, minimally intrusive code modifications, our technique can be readily integrated into existing pipeline implementations.
%
For the memory footprint of each cached tile per device, we formulate $M_{\text{tile}}$ in Equation~\eqref{eq:estimate-cache-memory}, with a batch size $B$, latent channel $C$, frame count $F$, and latent spatial $H \times W$:
\begin{equation} \small
M_{\text{tile}} = \underbrace{B \times C \times F \times H \times W}_{\text{Total Latent Elements}} \times \underbrace{2}_{\text{Bytes (bf16)}}.
\label{eq:estimate-cache-memory}
\end{equation}
% \begin{table}[t]
% \centering
% \caption{Communication Analysis}
% \label{tab:comm-cost}
% % 如果需要缩放，保留 resizebox，否则可以注释掉
% \resizebox{.7\linewidth}{!}{
%     \begin{tabular}{ccc}
%     \toprule
%     \multirow{2}{*}{\textbf{Method}} & \multicolumn{2}{c}{\textbf{Communication}} \\
%     \cmidrule(lr){2-3} % 这里只画在第2到第3列下面
%      & \textbf{Cost} & \textbf{Overlap} \\
%     \midrule
%     Tensor Parallel & $4O(p \times hs)L$           & \xmark \\
%     DistriFusion    & $2O(p \times hs)L$           & \cmark \\
%     SP-Ring         & $2O(p \times hs)L$           & \cmark \\
%     SP-Ulysses      & $\frac{4}{N}O(p \times hs)L$ & \xmark \\
%     PipeFusion      & $2O(p \times hs)$            & \cmark \\
%     Our Tile Parallel & $O(p)$                     & \xmark \\
%     \bottomrule
%     \end{tabular}
% }
% \end{table}

\begin{table}[t]
\centering
\caption{Comparison between different parallel methods
for DiTs in a single diffusion timestep. Overlap denotes the
overlapping between communication and computation. Notations: $p$: \#pixels tokens; $hs$: Model hidden size; $L$: \#model layers.}
\label{tab:comm-cost}
% resizebox 让表格自动适应宽度，这里设为 \linewidth 撑满
\resizebox{\linewidth}{!}{
    % 关键点：定义两组列，中间用 @{\hspace{...}} 强制插入间距
    % lcc (左边一组) + 间隔 + lcc (右边一组)
    \begin{tabular}{l|cc @{\hspace{0.1cm}} |l|cc}
    \toprule
    % --- 表头第一行 ---
    \multirow{2}{*}{\textbf{Method}} & \multicolumn{2}{c|}{\textbf{Communication}} & 
    \multirow{2}{*}{\textbf{Method}} & \multicolumn{2}{c}{\textbf{Communication}} \\
    
    % --- 表头下划线 (注意中间断开) ---
    \cmidrule(r){2-3} \cmidrule(l){5-6}
    
    % --- 表头第二行 ---
     & \textbf{Cost} & \textbf{Overlap} & & \textbf{Cost} & \textbf{Overlap} \\
    \midrule
    
    % --- 第 1 行数据 (左: Tensor, 右: Distri) ---
    Tensor Parallel & $4O(p \times hs)L$ & \xmark & 
    DistriFusion    & $2O(p \times hs)L$ & \cmark \\
    
    % --- 第 2 行数据 (左: SP-Ring, 右: SP-Ulysses) ---
    SP-Ring         & $2O(p \times hs)L$ & \cmark & 
    SP-Ulysses      & $\frac{4}{N}O(p \times hs)L$ & \xmark \\
    
    % --- 第 3 行数据 (左: PipeFusion, 右: Our Tile) ---
    PipeFusion      & $2O(p \times hs)$  & \cmark & 
    \cellcolor{gray!20}Tile Parallel & \cellcolor{gray!20}$O(p)$           & \xmark \\
    
    \bottomrule
    \end{tabular}
}
\end{table}
% The storage requirement scales linearly with sequence length $\mathcal{O}(F \cdot H \cdot W)$, with respect to the video length and resolution. 
For typical video generation tasks on HunyuanVideo ($C=32$, $F=80$, $720$p, VAE compression ratio = (4,8,8)), the total cached latent tile $M_{\text{tile}}$ remains about $100$ MB, which is negligible compared to the available VRAM on modern GPUs and model weights. Table~\ref{tab:comm-cost} shows the comparison of different parallelism methods~\cite{fang2024xditinferenceenginediffusion}. Since we communicate only latents, the cost of our Tile Parallelism is only $O(p)$, much smaller compared to Sequence Parallel~\cite{liu2023ringattentionblockwisetransformers, jacobs2023deepspeedulyssesoptimizationsenabling} and PipeFusion~\cite{wang2024pipefusion}. Therefore, the \texttt{allgather} incurs minimal latency, only about tens of milliseconds in our experiment (shown in \S\ref{sec:cache and parallel exp}). Consequently, our design maintains high throughput even as the number of devices or the video length increases.

\subsection{Cache-guided Workload Rebalance}

When the workload is imbalanced, such as with 9 tiles distributed across 4 GPUs, the key is to alleviate the load on the most heavily stressed device. Fortunately, our caching technique naturally facilitates this by identifying certain tiles as skippable at each step via \textit{Cache Predictor}, thereby creating opportunities to rebalance the workload.

After identifying local skippable tiles, we trigger the \textit{Tile Rebalancer} to perform an additional collective communication to exchange cache-related information about all tiles across all devices, including recent inputs and outputs required by our cache algorithm as well as metrics for our region-aware cache policy in \S\ref{subsec: Intra-step Region-aware Cache Control}. With this global information, each rank independently calculates a new, balanced workload distribution, thereby alleviating the load on the most heavily stressed device, as illustrated by the change in active tiles for GPU-3 before and after rebalancing in Figure~\ref{fig:sec4_4_tile_parallellism}. 
% \mlee{looking at fig 10, rebalancing with the least comm overhead seems to be swapping tile 1 and tile 6 between GPU0 and GPU3. is this local decision of each rank comm-efficient?}\fanjiang{now we simply use all gather to reorder all the tiles, not deciding the most optimal swapping.} 
Furthermore, by coalescing multiple tensors into a single \texttt{allgather} call, the communication cost becomes negligible relative to the computational cost. This efficient communication pattern underpins the excellent and strong scalability of our tile parallelism.

%%%%%%%%%%%%%%%%%%%%%%%%%%%%%%%%%%%%%%%%%%%%%%%%%%%%%%%%%%%%%%%%%%%%%%%%%%%%%%%%%%%%%%%%%
% Section 7 Programming support
% \section{\Mname{} Programming Support}
% \fanjiang{\Mname{} follows a multi-GPU design combined with caching and re-balancing, and compatible with different custom diffusion models. Listing~\ref{lst:listing-python} shows example for the tile-based video generation. There are three major components: 1) \textit{Distributed Manager} (Line 7) manages configurations, tensors, data movement, and communications, ensuring readability and maintainability, and employs \texttt{get\_boundary} to track each tile’s status and position. 2) \textit{CachingTransformer} (Line 8) exploits temporal and spatial redundancy via our caching mechanism. 3) \textit{Rebalancing} (Line 19) utilizes \texttt{allgather} to reorganize the tile-GPU configurations settings. Users will implement the \texttt{Tile\_gen} for running and easily extend it to custom diffusion models.}
% \input{listing/code_example}
% \section{Implementation}
% We implement \Mname{} with about 11k lines of Python code: 5k LOC inherited from HuggingFace Diffusers~\cite{von-platen-etal-2022-diffusers} and 6k LOC for adaptation. We extend two mainstream video generation models, CogVideoX~\cite{hong2022cogvideo} and HunyuanVideo~\cite{kong2024hunyuanvideo}, for evaluations, with PyTorch~\cite{paszke2019pytorch} as the main toolkit for tensor operations and inter-GPU communications.

\section{Implementation} 
\label{sec: implementation}
We implement \Mname{} with about 11k lines of Python code, including about 5k LOC inherited from HuggingFace Diffusers~\cite{von-platen-etal-2022-diffusers}, and another 6k LOC adapting the pipelines. We extended two mainstream video generation models for evaluations, CogVideoX~\cite{hong2022cogvideo} and HunyuanVideo~\cite{kong2024hunyuanvideo}. PyTorch~\cite{paszke2019pytorch} serves as the main toolkit for all tensor operations and inter-GPU communications.

During pixel-space upsampling (phase 2 in Figure~\ref{fig: sec4_2_TVG_framework}), we use the \textit{bicubic} interpolation algorithm and re-noise the latents for $k$ steps, with $k=45$ balancing quality and efficiency. 
% For tile-aware denoising, we employ \textit{TileTensor2D} to track each tile’s status and position. 
To exploit temporal and spatial redundancy, we subclass the transformer into \textit{CachingTransformer} and integrate a \textit{TileStdTracker} that computes per-tile noise standard deviations online and adapts region-aware cache thresholds by tracking related information needed by cache policy. For portability, we wrap all functionality within \textit{DistributedManager}, which manages configurations, tensors, data movement, and communications.
% , ensuring readability and maintainability
For cross-device communications, we use Pytorch \texttt{Allgather} primitive with NVIDIA NCCL~\cite{nccl2024}.

% \paragraph{Video Quality} To evaluate the quality of generated videos with our tiling-based scaling method, we choose the Vbench ~\cite{} as the quality evaluation benchmark. Vbench has about 700 prompts focusing on different aspect of the videos. We randomly pick about 500 prompts to generate different high-resolution videos.
% \paragraph{E2E Performance} 
% To learn about the efficient accelerating method in our design, we record the generation time w/ and w/o our hierarchical caching mechanism.  

\section{Evaluation}
This section presents our evaluation settings and findings. It is organized into four parts: (1) video quality benchmarks across different models and resolutions, (2) end-to-end performance, (3) effect of cache mechanism and scalability of tile parallelism, and (4) ablation studies on key hyperparameters. 
% This organization highlights the quality, performance, scalability, and effectiveness of \Mname{}. 

\begin{table}[t]
  \centering
  \caption{ Quality results of \Mname{} on VBench benchmark. V1–V5 denote the five evaluation metrics: \textbf{V1}: Subject Consistency, 
  \textbf{V2}: Background Consistency, \textbf{V3}: Motion Smoothness, 
  \textbf{V4}: Aesthetic Quality, and \textbf{V5}: Imaging Quality.}
  \label{tab:quality_benchmark}
  \resizebox{\linewidth}{!}{
\begin{tabular}{c|c|cccccc}
    \toprule
    Model & Setting & V1(\%) & V2(\%) & V3(\%) & V4(\%) & V5(\%) & Avg. \\
    \midrule
    \multirow{5}{*}{Cogvideo}
      & 720p          & 96.29 & 96.23 & 98.41 & 61.88 & 70.20 & \cellcolor{gray!20}\textbf{84.60} \\
      & 2K w/o Cache  & 95.66 & 96.06 & 97.22 & 63.86 & 70.38 & \cellcolor{gray!20}\textbf{84.64} \\
      & 2K w/ Cache   & 95.45 & 95.91 & 97.21 & 62.75 & 69.75 & \cellcolor{gray!20}\textbf{84.21} \\
      & 4K w/o Cache  & 92.94 & 94.11 & 98.10 & 57.92 & 67.38 & \cellcolor{gray!20}\textbf{82.09} \\
      & 4K w/ Cache   & 93.22    & 94.32   & 98.04    & 57.95    & 67.56    & \cellcolor{gray!20}\textbf{82.22} \\
    \midrule
    \multirow{5}{*}{Hunyuanvideo}
      & 720p          & 98.55 & 97.86 & 99.53 & 64.54 & 70.83 & \cellcolor{gray!20}\textbf{86.26} \\
      & 2K w/o Cache  & 98.02 & 97.31 & 99.42 & 66.47 & 69.62 & \cellcolor{gray!20}\textbf{86.17} \\
      & 2K w/ Cache   & 98.30 & 97.48 & 99.44 & 66.16 & 70.26 & \cellcolor{gray!20}\textbf{86.33} \\
      & 4K w/o Cache  & 97.76 & 97.24 & 99.43 & 63.07 & 69.68 & \cellcolor{gray!20}\textbf{{85.44}} \\
      & 4K w/ Cache   & 98.12 & 97.58 & 99.51 & 62.57 & 70.31 & \cellcolor{gray!20}\textbf{85.62} \\
    \bottomrule
\end{tabular}
  }
  \vspace{-10pt}
\end{table}

% \noindent 
\textbf{Setup.} We evaluate \Mname{} on a single node with eight NVIDIA H100-80GB GPUs interconnected via NVLink~\cite{nvlink2016} and two AMD EPYC 9534 64-Core CPUs. The environment includes Ubuntu 22.04.5 LTS, CUDA 12.2, PyTorch 2.5.1, and Diffusers 0.33.1.
We integrate CogVideoX-1.5 (5B)~\cite{hong2022cogvideo} and HunyuanVideo (13B)~\cite{kong2024hunyuanvideo} into our \Mname{} framework. 
% \mlee{integrating a framework into ML models sounds weird. you can say "integrate these models into our framework"}
Table~\ref{tab:video_model_perf} 
reports each model’s maximum supported resolution. 

% \noindent 
\textbf{Baselines.} To the best of our knowledge, there is no other training-free ultra-high-resolution video generation framework. To show the superiority of our caching mechanism, we choose the current state-of-the-art baseline TeaCache~\cite{liu2024timestep} for comparison, and implement it on our \Mname{} framework. To evaluate the benefits of our tile parallelism, we select xDiT~\cite{fang2024xditinferenceenginediffusion} as the baseline.

% \noindent 
\textbf{Metrics.} We evaluate \Mname{} on image-to-video (I2V) tasks using the VBench quality score~\cite{huang2023vbench} (covering subject consistency, background consistency, motion smoothness, aesthetic quality, and imaging quality), end-to-end latency and speedup from caching, and multi-GPU scalability. To characterize the quality-efficiency trade‑off introduced by caching, we compare generated videos w/ and w/o caching in terms of PSNR~\cite{wang2004image}, SSIM~\cite{wang2004image}, and LPIPS~\cite{zhang2018unreasonable}.

% \noindent 
\textbf{Workloads.} For quality evaluation, we randomly choose representative prompts including different amounts of dynamicity and scenes from VBench~\cite{huang2023vbench} to generate 2K and 4K videos for all models. For end-to-end performance, we run on all models with optimal cache thresholds and scaling factors. For multi-GPU scalability, we execute CogVideoX and HunyuanVideo on 1, 2, 4, and 8 GPUs and measure latency. For ablations, we vary shifting steps, tile numbers, and shifting frequency, with CogVideoX on 2K resolution.
% \paragraph{Baselines}
% To the best of our knowledge, no prior work offers training‑free, arbitrary‑resolution scaling for DiT‑based video generation. For comparing caching mechanisms, we adopt TeaCache~\cite{liu2024timestep} as our baseline, which is the current SOTA cache methodology. 

Unless stated otherwise, we set the \textit{loop step} to 16 (shift stride $=1/16$ tile size), forcing shift every step, and use 160$\times$90 (latent space) as the default tile size. All videos are 40 frames, 5s long. Basic cache thresholds are 0.09 (CogVideo) and 0.05 (HunyuanVideo).
% \mlee{I thought cache threshold is adaptive (i.e., dynamically updated) in sec 5 (e.g. 5.2). Here why using a fixed value?}\fanjiang{FJ: Here we adaptively modify the thres based on a basic thres.} 
With 720p tile as the base, 2K is split into 4 tiles and 4K into 9 tiles.

\begin{figure}[t]
  \begin{center}
  \includegraphics[width=\linewidth]{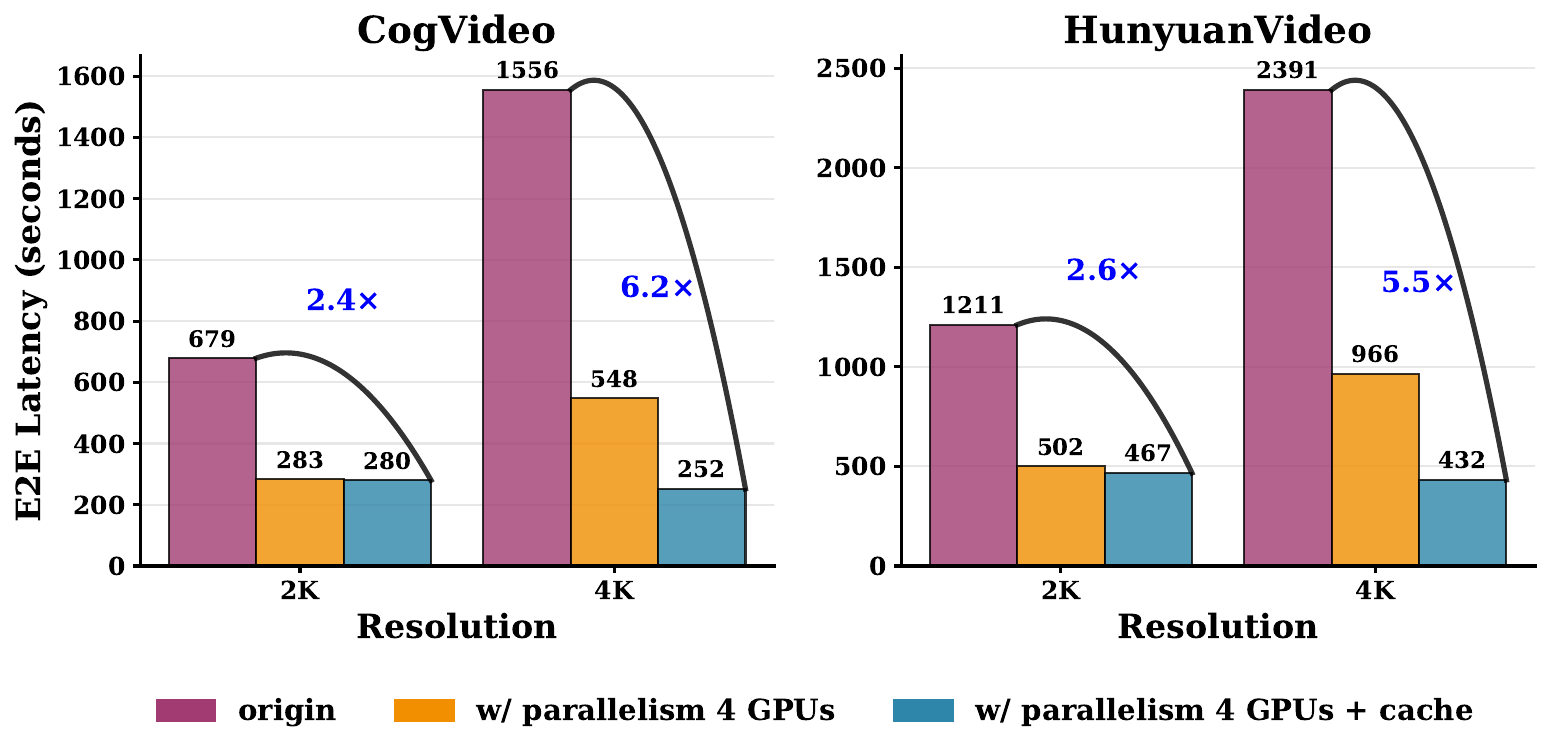}
  \end{center}
  \vspace{-10pt}
  \caption{End-to-end latency. Origin setting is evaluated with 1 GPU without cache. The other two settings with parallelism are measured with workload rebalance.}
  \label{fig: e2e_performance_comparison}
\end{figure}

\subsection{End-to-End Performance}
\begin{figure}[t]
  \begin{center}
  \includegraphics[width=\linewidth]{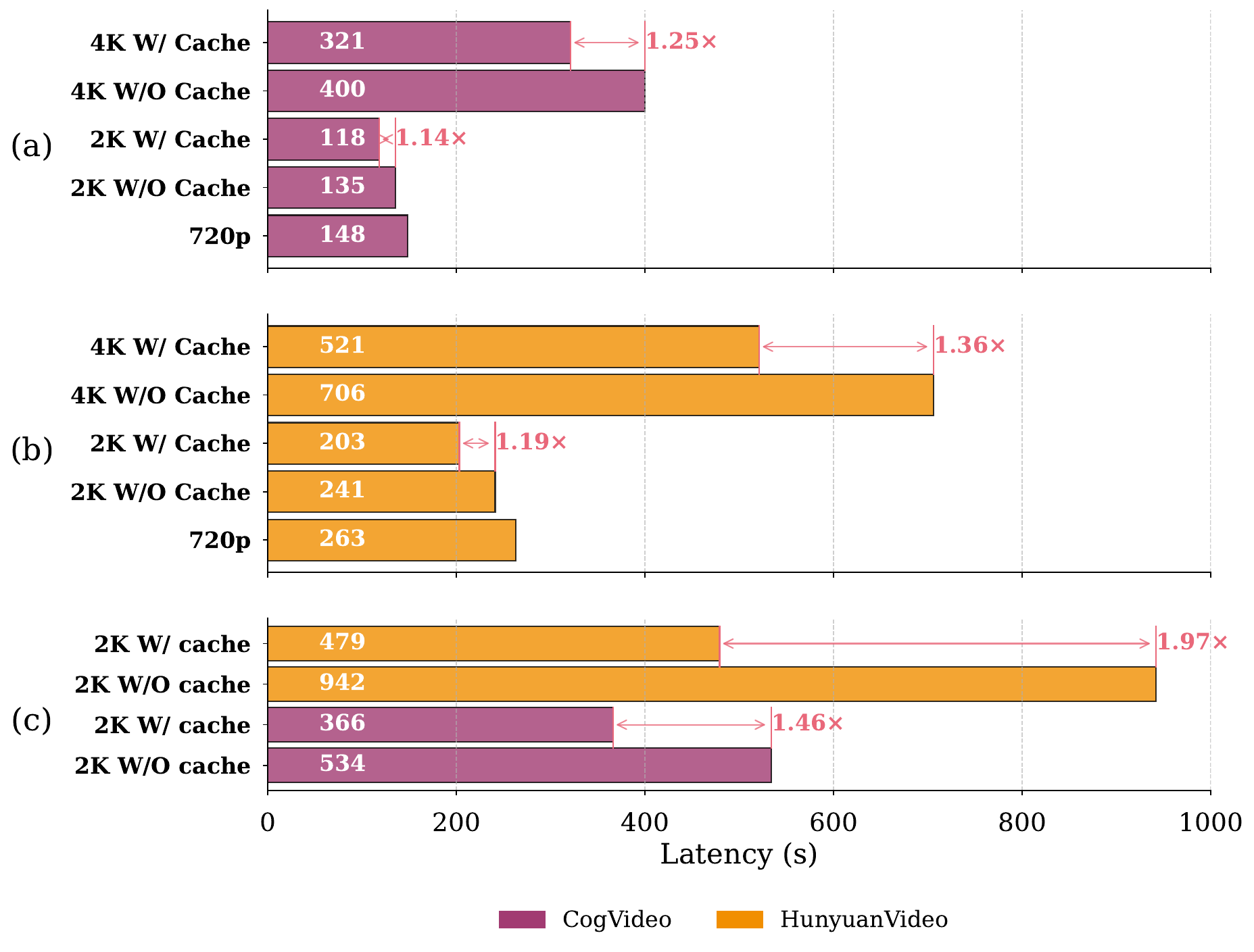}
  \end{center}
  \vspace{-10pt}
  \caption{Latency comparison between w/ and w/o cache. \textbf{(a)} and \textbf{(b)} are tested with 4 GPUs, and \textbf{(c)} with 1 GPU. For all settings, workload rebalance is disabled.}
  \label{fig:cache_latency_comparison}
\end{figure}
\textbf{Video quality.}
We consider three output resolutions: 720p,
% (the first-stage output in Figure~\ref{fig: sec4_2_TVG_framework}), 
2K, and 4K. 
% We select five dimensions from VBench++~\cite{huang2024vbench++} quality benchmark, spanning from temporal to frame-wise quality. 
Following the guidance in VBench++~\cite{huang2024vbench++}, we randomly sample 30 prompts for 2K and 20 prompts for 4K in multiple categories, preserving enough representivity to different types of videos. 
Table~\ref{tab:quality_benchmark} shows that \Mname{} attains nearly identical scores across 720p, 2K, and 4K, indicating it retains quality while upscaling and caching incurs negligible loss.  
When cache is enabled, we sometimes see an increase in VBench scores. 
This effect likely stems from the diffusion process and model properties: training on discrete timesteps can cause non-uniform or overfitting performance across the noise schedule. In such cases, reusing cached activations can accidentally compensate for weaknesses in the model, thereby improving sample quality.

\label{sec:8.2-e2e-performance}

% \vspace{-5pt}
\noindent\textbf{Runtime Performance.} Figure~\ref{fig: e2e_performance_comparison} reports the end-to-end (E2E) running time under three settings. 
% E2E time includes both the first stage and the second stage. 
With parallelism and no cache, \Mname{} achieves 2.5$\times$ to 2.8$\times$ speedup over the original settings. With cache enabled, the speedup comes to 5.5$\times$ and 6.2$\times$ on 4K resolution. This indicates both the scalability of our parallelism and the benefits of the cache mechanism. For 2K resolution, enabling cache does not provide a significant improvement in latency. This is because we split 2K videos into 4 tiles, and they could be evenly distributed across 4 GPUs. For each denoising step, only when 4 tiles are skipped at the same time can it reduce the overall latency. On the other hand, 
% \mlee{side $\rightarrow$ hand(?)} 
4K greatly benefits from cache-guided workload rebalance. As long as at least one tile 
% \mlee{minor: 1, 2 and 3 $\rightarrow$ one, two, and three. so here, one tile} 
is cached at a step, by the workload rebalance mechanism, \Mname{} reduces the longest workflow path from 3 tiles to 2 tiles, thus remarkably reducing latency. To highlight the benefits of our contribution, the first stage currently does not fully utilize all four GPUs, but its performance can be further improved by applying standard parallelization techniques such as xDiT~\cite{fang2024xditinferenceenginediffusion} to the first stage, which would yield an even lower E2E latency (in Figure~\ref{fig: breakdown_stages}). Moreover, we report the breakdown of E2E inference time on HunyuanVideo to analyze each optimization component (Figure~\ref{fig: techniques_breakdown}). Each design contributes significantly, achieving a total of $5.53\times$ speedup.
\subsection{Cache Efficiency and Parallel Scalability} 
\label{sec:cache and parallel exp}

\begin{figure}[t]
  \begin{center}
  \includegraphics[width=\linewidth]{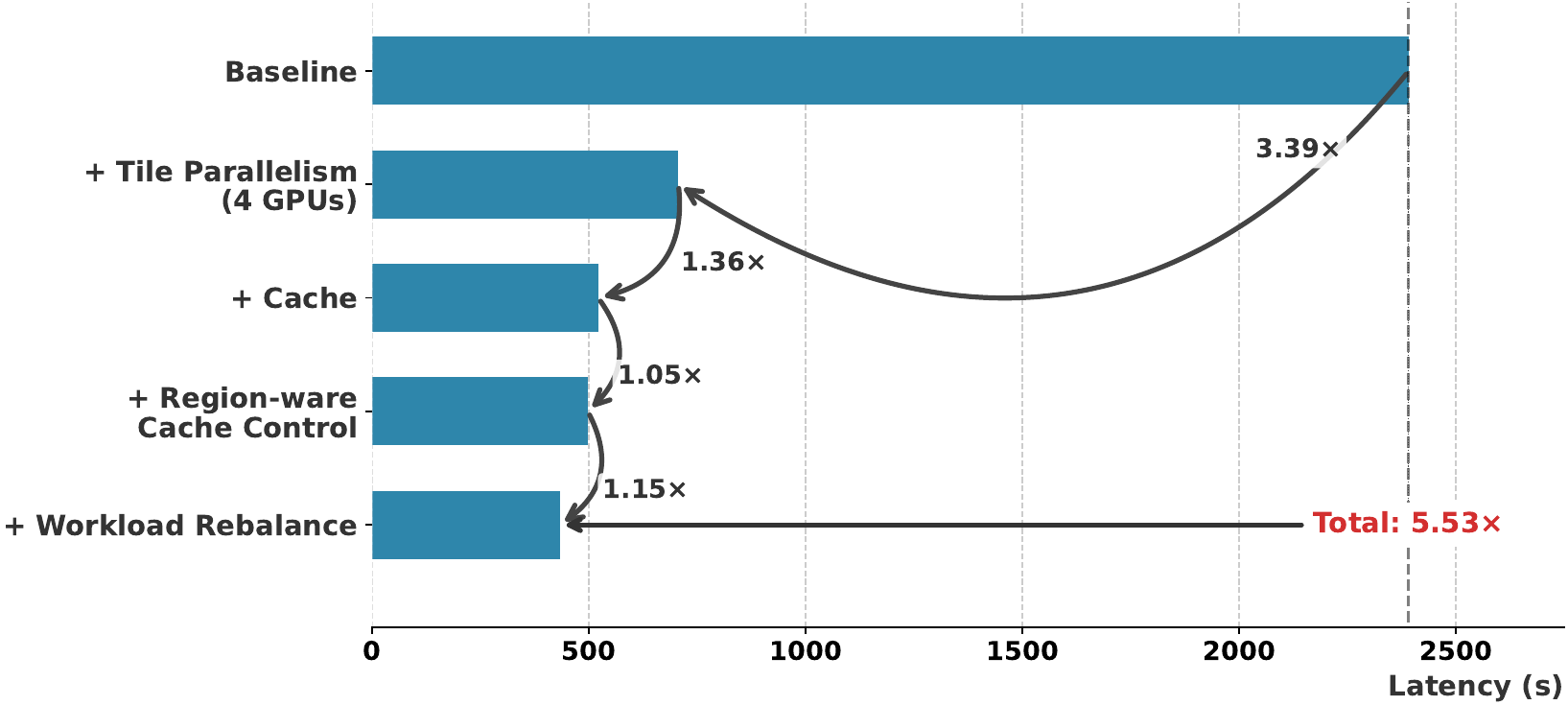}
  \end{center}
  \vspace{-10pt}
  \caption{The Breakdown of end-to-end runtime of HunyuanVideo when generating a 5.3s long 4K-resolution video. \Mname{} effectively reduces the end-to-end inference time from 2391 seconds to 432 seconds through system-algorithm co-design. Each design point contributes to a considerable improvement, with a total $5.53\times$ speedup.}
  \label{fig: techniques_breakdown}
\end{figure}

\begin{table}[t]
\centering
\caption{Speedup and quality metrics were tested w/ our cache enabled. The latency column is arranged as with/without caching. Visual Retention scores are computed by measuring the similarity of the w/ cache output to the w/o cache output (ground truth).}
\label{tab:cache speedup}
% \vspace{-5pt}
\resizebox{\linewidth}{!}{
\begin{tabular}{l|cc|ccc|c}
\toprule

                             & \multicolumn{2}{c|}{Latency}                                  & \multicolumn{3}{c|}{Visual Retention}                                    &                                         \\
\cellcolor[HTML]{DAE8FC}2K(1 GPU)   & Latency (s)$\downarrow$ & \multicolumn{1}{c|}{Speedup$\uparrow$} & PSNR$\uparrow$ & SSIM$\uparrow$ & \multicolumn{1}{c|}{LPIPS$\downarrow$} & \multirow{-2}{*}{VBench (\%)$\uparrow$} \\ \hline
CogVideo                     & 366 / 534                     & 1.46$\times$                           & 20.81             & 0.76            & 0.18                                     & 84.57                                      \\
Hunyuan & 479 / 942                     & 1.97$\times$                           &  36.61            &  0.96            &   0.02                                   & 86.11                                      \\ \hline
\cellcolor[HTML]{DAE8FC}2K(4 GPU)    & \multicolumn{6}{c}{} \\ \hline
CogVideo                     & 118 / 135                     & 1.14$\times$                           & 20.62             & 0.77            & 0.17                                     & 84.21                                      \\
Hunyuan & 203 / 241                     & 1.19$\times$                           &  34.87            &  0.93            &   0.04                                   & 86.33                                      \\ \hline

\cellcolor[HTML]{FFCE93}4K(4 GPU)   & \multicolumn{6}{c}{}                                                                                                                                                                  \\ \hline
CogVideo                     & 321 / 400                     & 1.25$\times$                           & 23.22             & 0.84             & 0.17                                     & 82.22                                      \\
Hunyuan & 521 / 706                     & 1.36$\times$                           & 34.73             & 0.91             & 0.07                                     & 85.62    \\  \bottomrule
\end{tabular}
}
\end{table}

\begin{table}[t]
\centering
\caption{Comparison of speedup and quality metrics w/ different cache mechanisms on 2K resolution, tested on the single GPU. Visual Retention is to compare w/o cache (ground truth) and w/ cache.}
\label{tab:cache speedup comparison}
\resizebox{\linewidth}{!}{
\begin{tabular}{l|cc|ccc}
\toprule
% Use multirow{2} here to span down two rows.
% Removed extra &'s so the column counts match the definition.
\multirow{2}{*}{Settings} & \multicolumn{2}{c|}{Efficiency} & \multicolumn{3}{c}{Visual Retention} \\
 & Latency (s)$\downarrow$ & Speedup$\uparrow$ & PSNR$\uparrow$ & SSIM$\uparrow$ & LPIPS$\downarrow$ \\ \hline
Original & 604 & - & - & - & - \\ \hline
TeaCache-Slow & 536 & 1.13$\times$ & 14.97 & 0.72 & 0.21 \\
TeaCache-Slow & 412 & 1.47$\times$ & 14.87 & 0.71 & 0.21 \\ 
Our-Slow & 322 & 1.88$\times$ & 14.94 & 0.72 & 0.21 \\
Our-Fast & 296 & 2.04$\times$ & 14.81 & 0.70 & 0.24 \\ 
\bottomrule
\end{tabular}
}
\end{table}
\textbf{Cache.}
To evaluate the effectiveness of our caching mechanism, we compare the stage-2 latency and fidelity (PSNR, SSIM, LPIPS) with and without cache. As shown in Figure~\ref{fig:cache_latency_comparison}, \Mname{} achieves up to $1.19\times$ speedup at 2K and $1.36\times$ at 4K.
% , with larger 
% % \mlee{stronger $\rightarrow$ larger?} 
% gains at 4K due to workload imbalance (\S\ref{sec:8.2-e2e-performance}).
% Figure~\ref{fig:cache_latency_comparison}(c) shows the latency tested on a single GPU. 
When testing on a single GPU in Figure~\ref{fig:cache_latency_comparison}(c), without synchronization across devices, every effective caching contributes to decreasing the overall latency, thus bringing a higher 
% \mlee{better $\rightarrow$ higher?} 
speedup up to $1.97\times$ at 2K. 
Meanwhile, Table~\ref{tab:cache speedup} lists the metrics for quality evaluation with and without cache: PSNR and SSIM remain high and LPIPS stays low~\cite{zhao2025realtimevideogenerationpyramid}, indicating that our \Mname{} preserves high perceptual quality while providing substantial acceleration support. 

To better demonstrate the superiority of our caching mechanism, we compare it with the current state-of-the-art caching technique, TeaCache~\cite{liu2024timestep}, on CogVideoX-1.5. Following the setting of the official repo, we consider two configurations: \textit{Slow} and \textit{Fast}. \textit{Slow} corresponds to less aggressive skipping, while \textit{Fast} corresponds to more aggressive skipping. As shown in Table~\ref{tab:cache speedup comparison}, our caching mechanism is much faster than TeaCache while achieving similar or even better video quality. Moreover, our mechanism avoids the extensive offline profiling overhead required to obtain the approximation coefficients in~\cite{liu2024timestep}, making it much more convenient and more easily applicable to new diffusion models.

\noindent\textbf{Tile Parallelism.}
To assess the performance of our tile parallelism, we measure the stage-2 latency. As in Figure~\ref{fig:GPU_scalability_comparison},
\Mname{} shows linear scalability on 2K (2.0$\times$ on 2 GPUs, 4.0$\times$ on 4 GPUs) and sublinear on 4K (2.1$\times$, 3.5$\times$, and 4.4$\times$ on 2, 4, and 8 GPUs). 2K videos are not benchmarked beyond 4 GPUs because we split 2K videos into only 4 tiles.
% 2K tasks give better scalability because 4 tiles can always be evenly distributed across 2 and 4 GPUs, whereas 4K videos are split into 9 tiles and can never be evenly distributed, resulting in 1 device being more heavily loaded than others. 
The workload imbalance in 4K resolution generation (9 tiles) leads to sublinear scalability. We also compare tile parallelism with xDiT~\cite{fang2024xditinferenceenginediffusion} on CogvideoX with 45 diffusion steps. We apply xDiT on each tile and report the runtime with 1 to 8 GPUs. Table~\ref{tab:parallelism_speedup_comparison} shows that when the number of tiles is divisible by the number of GPUs, our tile paralleism can achieve better scalability. When it cannot be evenly split, our method can still achieve a similar or even better speedup. Furthermore, our method is orthogonal to xDiT and can be applied together on different dimensionalities.

\noindent\textbf{Workload Rebalance.} When cache is enabled, workload rebalance alleviates the imbalance between devices. Figure~\ref{fig: multimodel_multiGPU_latency_reallocate_comparison} depicts the speedup by workload rebalance on the basis of tile parallelism. Reducing GPU idle time caused by synchronization and workload imbalance, \Mname{} gains up to another 1.42$\times$ speedup with 8 GPUs on 4K video. Besides, for \texttt{allgather} primitives, we observe extremely low latency of about 10 ms, compared to hundreds of seconds E2E latency.

\begin{figure}[t]
  \begin{center}
  \includegraphics[width=\linewidth]{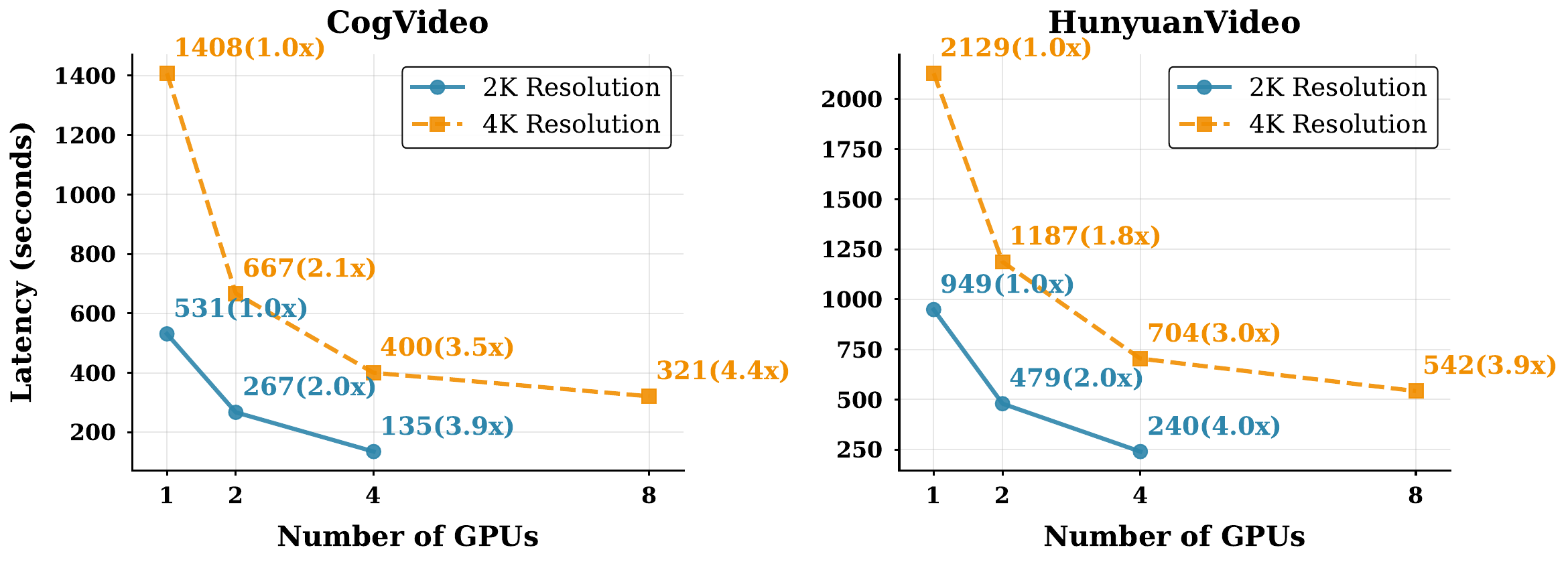}
  \end{center}
  \vspace{-10pt}
  \caption{Tile parallelism scalability. All settings are tested without cache and workload rebalance. Latency involves only the second stage.}
  \label{fig:GPU_scalability_comparison}
\end{figure}

\begin{figure}[t]
  \begin{center}
  \includegraphics[width=\linewidth]{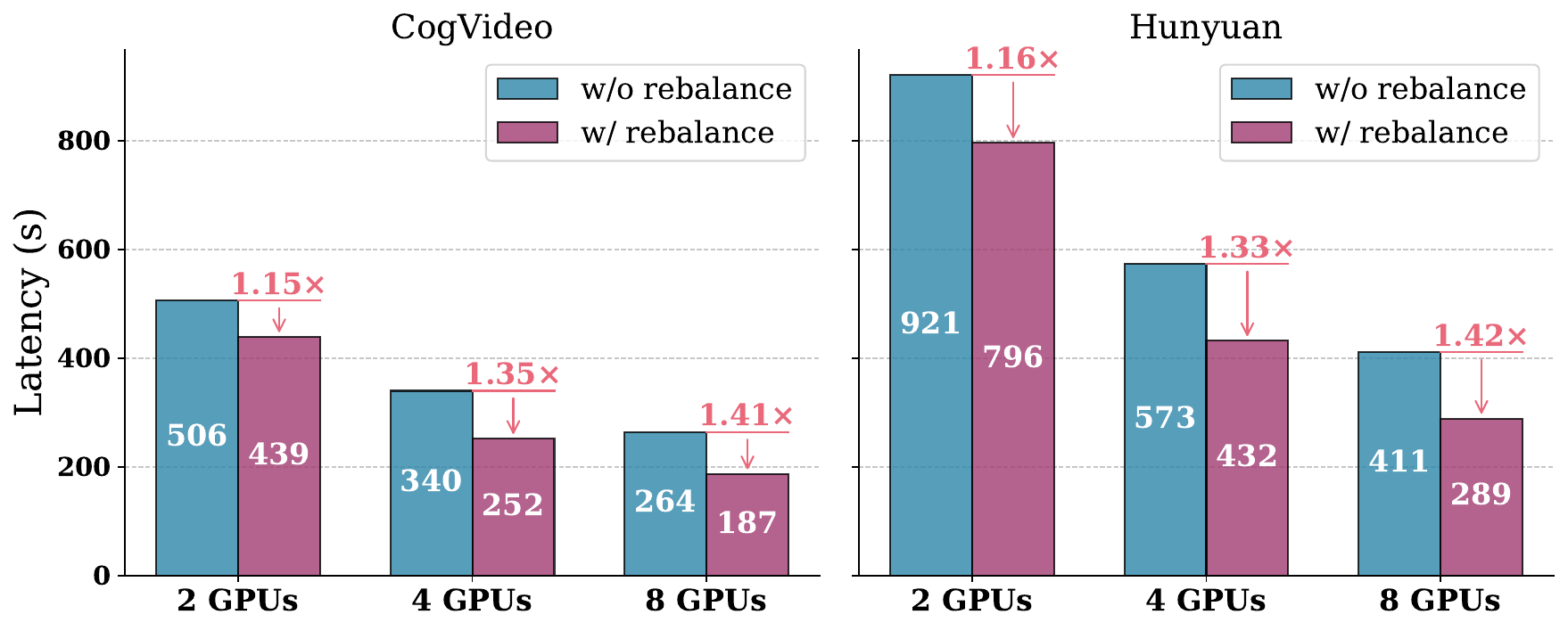}
  \end{center}
  \vspace{-10pt}
  \caption{Latency comparison between w/ and w/o reallocation. All settings are tested with 4K resolution and cache enabled. Latency involves only the second stage.}
  \label{fig: multimodel_multiGPU_latency_reallocate_comparison}
\end{figure}

\begin{table}[t]
\centering
\caption{Comparison of latency (seconds) between our tile parallelism and xDiT across different resolutions on CogVideoX.} 
\label{tab:parallelism_speedup_comparison}
\resizebox{\linewidth}{!}{
\begin{tabular}{lcccc} 
\toprule
Setting & 1 GPU & 2 GPUs & 4 GPUs & 8 GPUs \\
\midrule
xDiT-2K / 4K & 583 / 1312 & 334 / 750 & 194 / 437 & 126 / 284 \\
Tile Parallelism-2K / 4K & 534 / 1408 & 267 / 667 & 135 / 400 & - / 321 \\
\bottomrule
\end{tabular}
}
\end{table}

\subsection{Ablation Study}
\label{subsec: ablation study}
% In this part, we study various aspects of \Mname{} framework and evaluate the design choices we make with hyperparameters. Unless otherwise stated, CogVideo serves as the default subject model. 

% \begin{figure}[t]
%   \begin{center}
%   \includegraphics[width=\linewidth]{figures/Shift_stride_effect.pdf}
%   \end{center}
%   \caption{Shift stride effect.}
%   \label{fig: Shift_stride_effect}
% \end{figure}
\noindent\textbf{Tile-shifting impact.} We vary \emph{shift frequency} and \emph{shift stride} to study their effects.
It shows that no shifting gives the worst quality with seams and boundary inconsistency with VBench Score $81.56\%$ (detailed in Table~\ref{tab:quality_benchmark_abalation} at Appendix~\ref{appendix: Additional Ablation Studies}), while shifting more than once every 5 rounds shows no significant difference in VBench scores around $83\%$. It's worth noting that the VBench benchmark is better at recognizing global inconsistency than thin discontinuity at the tile boundary. Although barely perceptible, visual inspection confirms that frequent shifting stabilizes outputs. Since shifting incurs only negligible overhead, we demand shifting every round in all of our evaluation settings.
For shift stride, all settings give similar quality with VBench score around $83\%$, showing insensitivity of our techniques to this hyperparameter. Therefore, we choose a medium choice of 16 for all other settings.

\begin{figure}[t]
  \begin{center}
  \includegraphics[width=\linewidth]{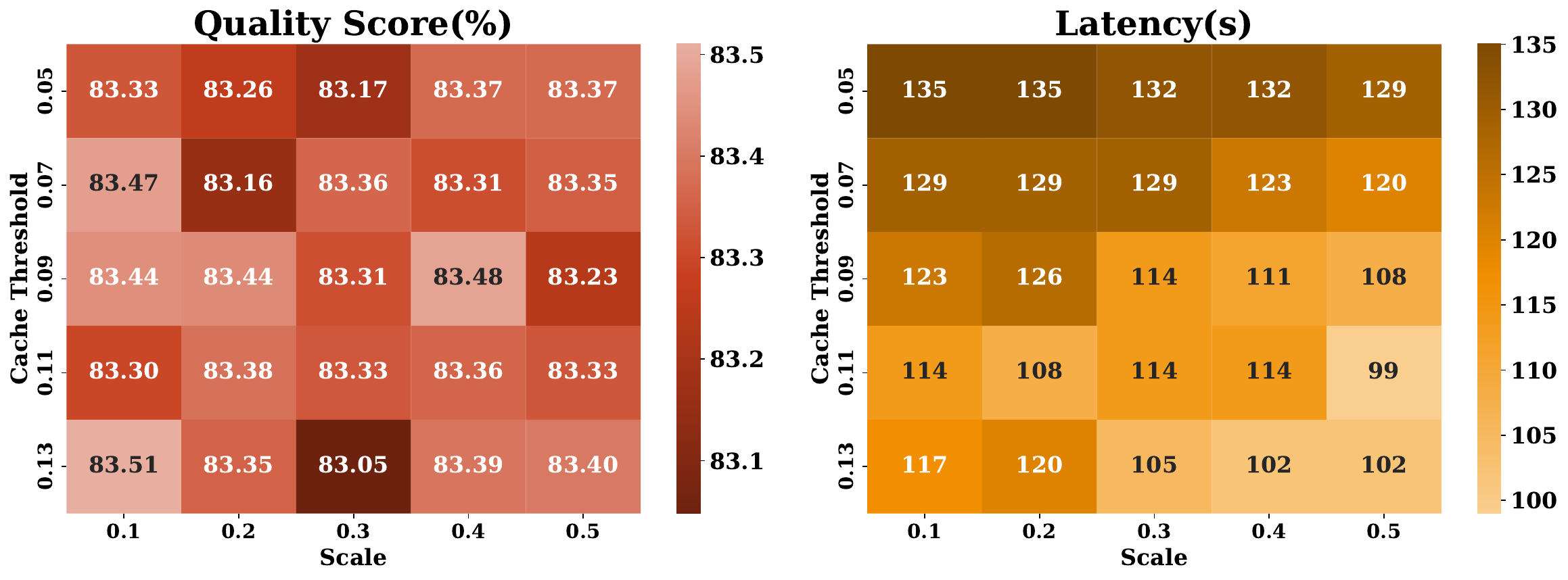}
  \end{center}
  \vspace{-5pt}
  \caption{VBench scores (left) and generation latency (right) measured with different cache thresholds and scale factors. Latency includes only the second stage time consumption with 4 GPUs. The resolution is 2K split into 4 tiles.}
  \label{fig: cache_threshold_effect}
\end{figure}

\noindent\textbf{Number of tiles.}
% DiT models rely on attention mechanism to predict noise, which has the computation complexity of $O(N^2)$ in nature, where $N$ is the number of tokens, proportional to the number of pixels and frames. Also, the smaller tiles are, the more global information is lost to them during processing. As denoising goes on, this error will be accumulated across iterations and degrades the consistency in final output. Therefore, the size of tiles is of great significance to both latency and video quality. To depict this point, 
Figure~\ref{fig: num_of_tiles_effect} shows the efficiency–quality trade-off when varying tile counts. Due to the compatibility of baseline models, the dimension sizes must be even numbers and cannot be too small. We tested the combinations listed in the table on the RHS. It shows that, with a fixed global size, as the tile size decreases (more tiles), the latency decreases due to the independence of each tile discussed in \S\ref{sec_6-1}, while the quality also drops because tile shifting cannot handle too many inconsistent boundaries at once. 
% Standard 720p videos are of latent size 160x90, which is best supported by our baseline models thus being chosen as our default tile size.
\begin{figure}[t]
  \begin{center}
  \includegraphics[width=\linewidth]{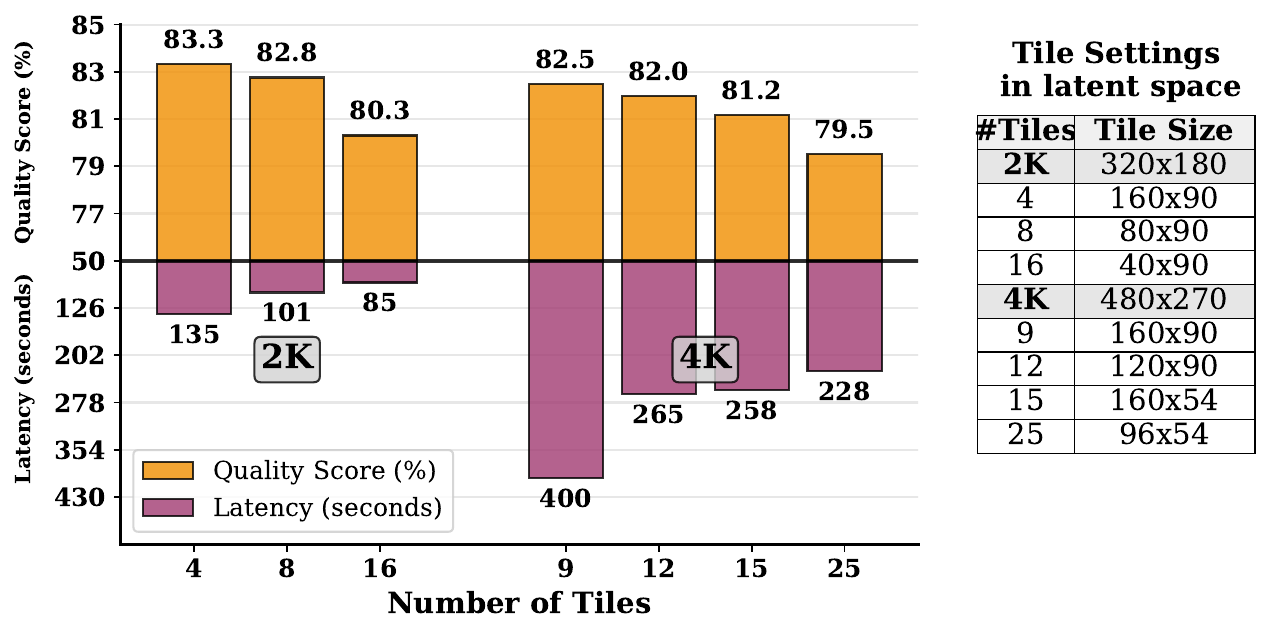}
  \end{center}
  % \vspace{-5pt}
  \caption{
  VBench scores and generation latency measured with different tile sizes. Quality score is the average of the five aspects, and latency includes only the second stage time consumption with 4 GPUs and cache disabled. Tile settings are listed at the RHS for reference.}
  \label{fig: num_of_tiles_effect}
\end{figure}

\begin{figure}[t]
  \begin{center}
  \includegraphics[width=\linewidth]{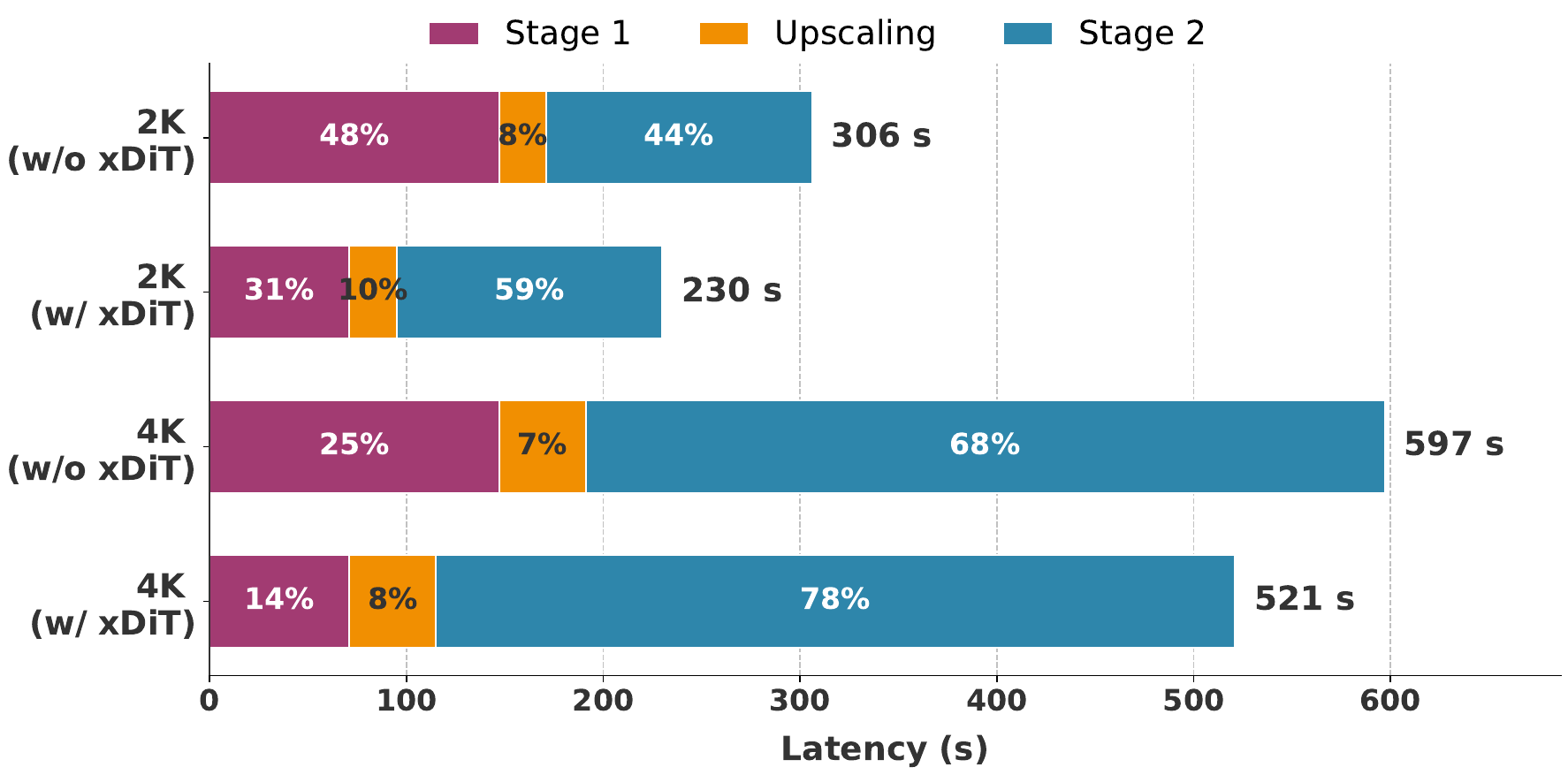}
  \end{center}
  \caption{
  Runtime breakdown across three stages on CogVideoX with 4 GPUs. Here w/ and w/o xDiT indicate whether xDiT is applied in the first-stage generation.}
  \label{fig: breakdown_stages}
\end{figure}

\noindent\textbf{Caching threshold.}
Figure~\ref{fig: cache_threshold_effect} shows quality and latency under varying \textit{cache threshold} and \textit{scale factor}. 
% A lower cache threshold makes more stringent control. A higher threshold allows more steps to be cached, thus resulting in more approximation error and less computation. The scale factor, on the other hand, further looses this control. A higher scale factor allows cache mechanism to more aggressively cache tiles with less dynamic content, further speedup denoising process. 
According to VBench score, different cache thresholds result in no significant difference in video quality. This is because our cache mechanism is good at preserving necessary information. 
% Although VBench cannot fully reveal the subtle changes that are visible to human observers. 
When we visually compare the videos from different thresholds, a higher threshold is more likely to produce an unstable view and blur some details. 
% For CogVideo, cache threshold 0.09 and scale factor 0.3 is chosen for other evaluations due to their balance between latency and quality.

\noindent\textbf{Analysis of Stages}. \Mname{} consists of three stages for video generation as illustrated in Figure~\ref{fig: sec4_2_TVG_framework}. We report the runtime breakdown of these three stages on CogVideoX with 4 GPUs in Figure~\ref{fig: breakdown_stages}. The results show that, whether applying parallelism in the stage-1 generation or not, stage-2 is the main bottleneck when generating 2K or 4K videos. Besides, to verify the necessity of stage-1 generation, we evaluate the quality of zero-shot output and \Mname{} output on VBench and a user study (detailed in Appendix~\ref{appendix: user study}) for fair comparison. As shown in Table~\ref{tab:zero-shot}, we found that Vbench does not fully recognize object duplication artifacts, while the user study shows a preference for the \Mname{} output w/ stage-1, highlighting the necessity of global semantic guidance.

\begin{table}[t]
\centering
\caption{Quantitative quality comparison of VBench scores and User Study ratings across 2K and 4K resolutions between zero-shot and \Mname{}.}
\label{tab:zero-shot}
\resizebox{.7\linewidth}{!}{
\begin{tabular}{lcccc}
\toprule
\multirow{2}{*}{Method} & \multicolumn{2}{c}{VBench ($\uparrow$)} & \multicolumn{2}{c}{User Study ($\uparrow$)} \\
\cmidrule(lr){2-3} \cmidrule(lr){4-5} 
 & 2K & 4K & 2K & 4K \\
\midrule
Zero-shot & 0.818 & 0.815 & 1.67 & 1.17 \\
Ours & 0.821 & 0.822 & 4.30 & 4.70 \\
\bottomrule
\end{tabular}
\vspace{-20pt}
}
\end{table}

\section{Related Work}
% System-level optimizations are crucial for deploying diffusion models efficiently, especially for ultra-high-resolution video generation. 
% These optimizations often focus on reducing the computational cost per step or improving throughput through parallelism and resource management. 

% Caching
\textbf{Caching Mechanism.}
Caching techniques reduce per-step computation in diffusion models by exploiting temporal locality and reusing intermediate results. 
Early works on U-Net architectures focused on caching approximate noise states or intermediate feature maps~\cite{agarwal2024approximate, ma2024deepcache}. 
Subsequent efforts for DiTs explored diverse strategies, including output reuse, block skipping, and residual reuse~\cite{ma2024learning, selvaraju2024fora, shenmd, chen2024delta}, 
as well as techniques such as reusing shared components between conditional and unconditional outputs~\cite{lv2024fastercache}, 
predicting token importance for selective computation~\cite{lou2024token}, alternating between full-feature and token-wise reuse~\cite{zou2024accelerating}, 
adapting policies based on residual changes or motion dynamics~\cite{kahatapitiya2024adaptive}, and
estimating output variations via timestep-modulated inputs~\cite{lv2024fastercache}. 
These approaches primarily target inter-step redundancy, whereas our method provides finer-grained control by exploiting both inter-step and intra-step reuse to capture spatio-temporal redundancy more effectively.

\noindent\textbf{Parallelization Mechanism.}
Diffusion inference is inherently computation-intensive. 
To mitigate this, prior works developed parallelism strategies such as Distrifusion to partition inputs into spatial patches~\cite{li2024distrifusion, zhang2024partially}, 
Asyncdiff to distribute model components to enable asynchronous execution~\cite{chen2024asyncdiff}, 
or PipeFusion to combine patch-level and pipeline-level partitioning to overlap computation and communication~\cite{wang2024pipefusion}. 
In contrast, we exploit an independent tile parallelism that minimizes synchronization and intensive communication cost, further incorporating cache-guided workload rebalancing to maximize the GPU utilization.

\section{Conclusion}
This paper introduces \Mname{}, a powerful framework to generate high-quality videos at ultra-high resolutions based on the original supported resolution. 
\Mname{} features a novel
training-free algorithmic innovation through effective sketch-tile collaboration.
To support such a new algorithmic design, \Mname{} combines a memory- and compute-efficient tile-based framework, a fine-grained adaptive region-aware caching strategy, and an intelligent cost-efficient tile parallelism, to significantly accelerate generation while maintaining quality. Experiments show substantial speedups over those state-of-the-art solutions. 
\Mname{} is fully open-source to facilitate the high-quality video generation ecosystem and infrastructure development.

\bibliographystyle{plain}
\bibliography{references}

% % --- START OF APPENDIX CONFIGURATION ---
\appendix       % This command changes section numbering from 1, 2, 3 to A, B, C
\clearpage        % Start a new page
\section*{Appendix} % This prints the unnumbered title "Appendix" at the top

\section{Algorithm for Region-aware Cache}
\label{appendix:algo-2}

In this section, we present Algorithm~\ref{alg:region-aware-cache}, which is the algorithm for the region-aware cache algorithm. This algorithm supplements the discussion in Section~\ref{subsec: Intra-step Region-aware Cache Control}.

\begin{algorithm}[h] \small
\caption{Intra-step Region-aware Cache Control}
\label{alg:region-aware-cache}
\begin{algorithmic}[1]
\Require Video latents $\mathbf{L}$, timesteps $\{t_1,\dots,t_S\}$, base threshold $\tau_{base}$, scaling $\alpha$, update interval $\Delta$, mode $\mathcal{M}$
\Ensure Denoised video latents $\hat{\mathbf{L}}$
\vspace{0.5em}
\State Initialize per-tile thresholds $\{\tau_i\} \gets \tau_{base}$
\For{$s = 1$ to $S$}
    \ForAll{tiles $i$}
        \State $L_i \gets \textsc{ExtractTile}(\mathbf{L}, i)$
        \State $\epsilon_i \gets \textsc{Model}(L_i, t_s, \tau_i)$
        \State $\textsc{StdTracker.Update}(i, \epsilon_i)$ \Comment{\textcolor{blue}{Online Track}}
    \EndFor
    \State $\epsilon \gets \textsc{Fuse}(\{\epsilon_i\})$, \quad $\mathbf{L} \gets \textsc{Denoise}(\mathbf{L}, \epsilon)$
    
    \If{$s \bmod \Delta = 0$}
        \State $\{\sigma_i\} \gets \textsc{StdTracker.Normalized}()$ \Comment{\textcolor{blue}{Normalize}}
        \State Initialize adjustment factor vector $\boldsymbol{\delta} \gets \mathbf{0}$
        
        \If{$\mathcal{M} = \text{``MaxMin''}$} \Comment{\textcolor{blue}{``MaxMin'' Policy}}
            \State $k_{\max} \gets \arg\max_i \sigma_i, \quad k_{\min} \gets \arg\min_i \sigma_i$
            \State $\delta_{k_{\max}} \gets \alpha, \quad \delta_{k_{\min}} \gets -0.5\alpha$
        \ElsIf{$\mathcal{M} = \text{``TopSelect''}$} \Comment{\textcolor{blue}{``TopSelect'' Policy}}
            \State $k_1, k_2 \gets \textsc{TopKIndices}(\{\sigma_i\}, 2)$
            \State $\delta_{k_1} \gets \alpha, \quad \delta_{k_2} \gets 0.5\alpha$
        \EndIf
        
        \State $\{\tau_i\} \gets \tau_{base} \cdot (\mathbf{1} + \boldsymbol{\delta})$ 
    \EndIf
\EndFor
\State \Return $\mathbf{L}$
\end{algorithmic}
\end{algorithm}

%%%%%%%%%%%%%%%%%%%%%%%%%%%%%%%%%%%%%%%%%%%
\section{Visualization of Claims}
\label{appendix: Visualization of Claims}
Figure~\ref{fig: sec_4_2_1} supplements the discussion in Section~\ref{sec: algorithm framework}. Figure~\ref{fig: sec_4_2_1} (a) and (b) compare the effects of upscaling in latent space and pixel space. Figure~\ref{fig: sec_4_2_1}(c) shows the object repetition artifacts when generating videos without stage-1 generation. Figure~\ref{fig: sec_4_2_2_tile_shifting effect} shows the benefits brought from tile shifting, which removes the seams across the tile boundary.

\begin{figure}[h] 
  \begin{center}
\includegraphics[width=1.0\linewidth]{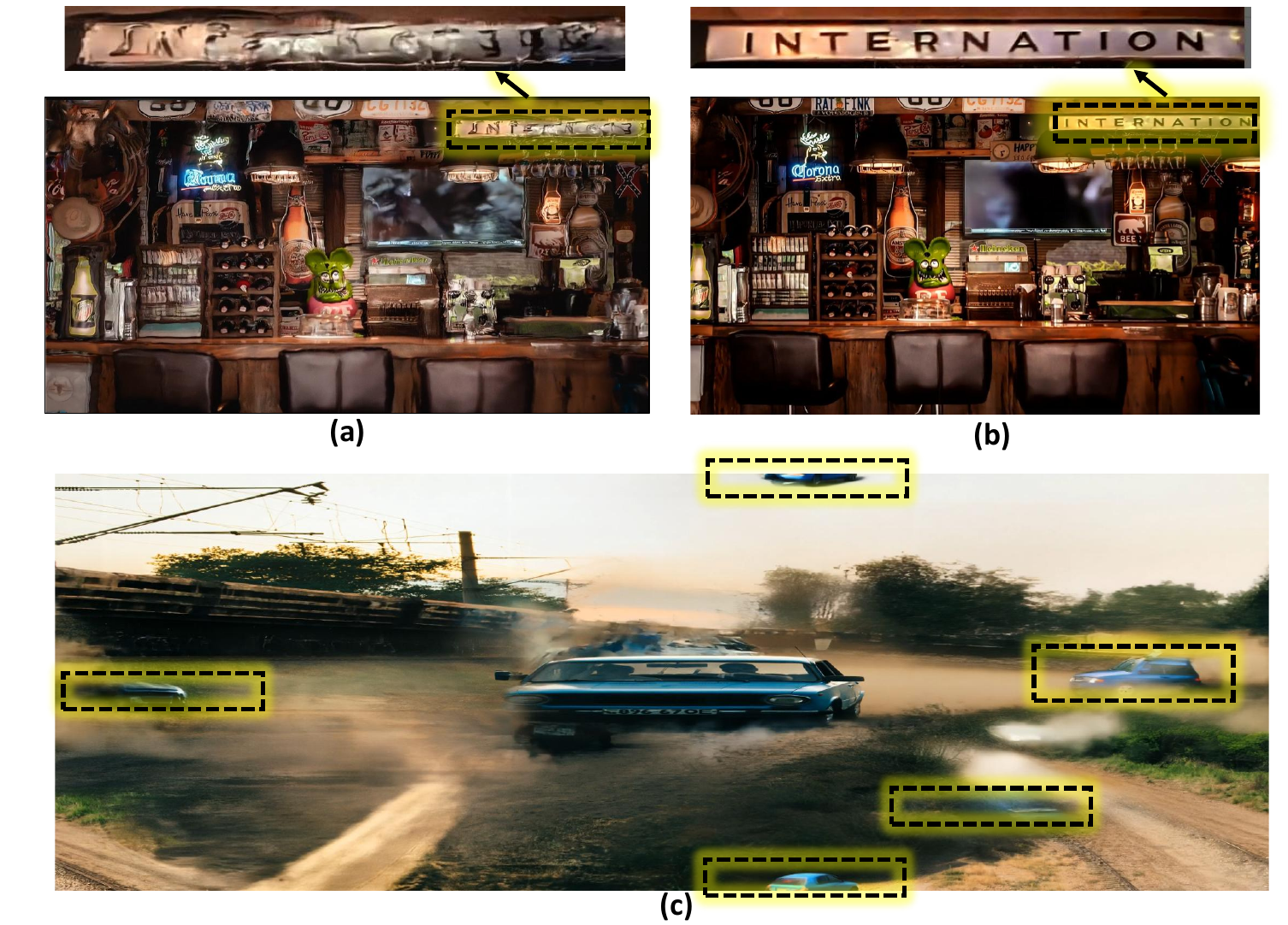}
  \end{center}
    \caption{(a) Interpolation in the latent space, which introduces noticeable artifacts and inconsistencies. (b) Interpolation in the pixel space, which yields more stable and visually consistent results. (c) Zero-shot generation without global semantic guidance, resulting in object duplication artifacts.
    }
  \label{fig: sec_4_2_1}
\end{figure}

\begin{figure}[h] 
  \begin{center}
\includegraphics[width=1.0\linewidth]{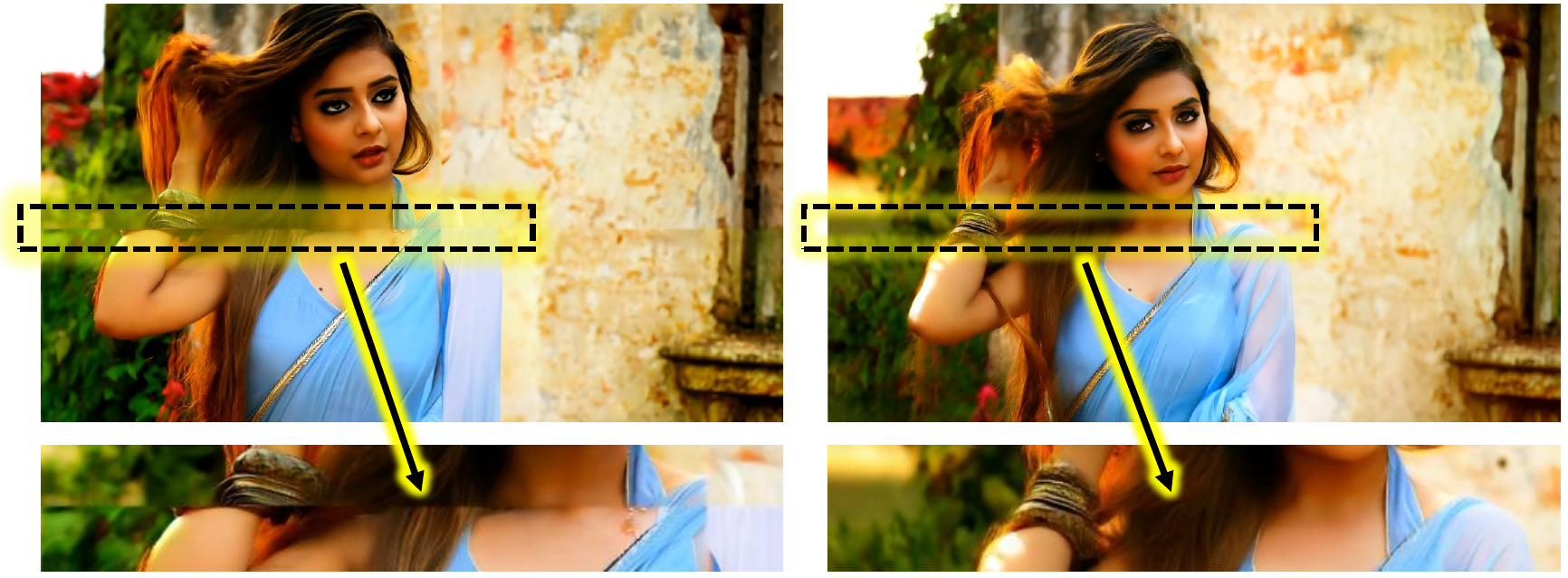}
  \end{center}
  \caption{Illustration of the effect of tile shifting. \textbf{Left}: Without tile shifting, noticeable inconsistencies appear along tile boundaries. \textbf{Right}: With tile shifting applied, the generated video exhibits seamless transitions and improved spatial consistency across tiles.
  }
  \label{fig: sec_4_2_2_tile_shifting effect}
\end{figure}

%%%%%%%%%%%%%%%%%%%%%%%%%%%%%%%%%%%
\section{Additional Ablation Studies}
\label{appendix: Additional Ablation Studies}
\begin{table}[h]
  \centering
  \caption{Quality evaluation results of \Mname{} on VBench scores.
  % V1–V5 denote the same five aspects as Table~\ref{tab:quality_benchmark}.
  \textit{loop step} parameter is defined as $(1/\mathit{shift}\_\mathit{stride})$. 
  The number $x$ in frequency test means shifting every $x$ steps.} 
  \label{tab:quality_benchmark_abalation}
  \resizebox{\linewidth}{!}{
\begin{tabular}{c|c|cccccc}
    \toprule
    Shift test & Setting  & V1(\%) & V2(\%) & V3(\%) & V4(\%) & V5(\%) & Avg. \\
    \midrule
    \multirow{6}{*}{Frequency}
      & no shift     & 92.72 & 93.91 & 97.98 & 54.76 & 68.45 & \textbf{81.56} \\
      & 15           & 92.92 & 94.19 & 98.06 & 59.66 & 67.81 & \textbf{82.53} \\
      & 9            & 92.91 & 94.43 & 98.09 & 61.93 & 66.97 & \textbf{82.87} \\
      & 5            & 93.26 & 94.77 & 98.14 & 62.36 & 68.51 & \textbf{83.41} \\
      & 3            & 93.01 & 94.42 & 98.16 & 62.14 & 67.92 & \textbf{83.13} \\
      & every step   & 93.16 & 94.19 & 98.22 & 62.20 & 67.73 & \textbf{83.10} \\
    \midrule
    \multirow{6}{*}{Loop step}
      & 4   & 93.37 & 94.67 & 98.23 & 63.09 & 67.25 & \textbf{83.52} \\
      & 8   & 93.39 & 94.93 & 98.23 & 62.89 & 68.06 & \textbf{83.50} \\
      & 16  & 92.95 & 94.89 & 98.15 & 62.52 & 67.26 & \textbf{83.35} \\
      & 32  & 93.11 & 94.35 & 98.20 & 62.76 & 68.67 & \textbf{83.42} \\
      & 64  & 92.91 & 94.28 & 98.15 & 62.22 & 67.77 & \textbf{83.07} \\
      & 90  & 93.13 & 94.50 & 98.15 & 61.43 & 67.93 & \textbf{83.03} \\
    \bottomrule
\end{tabular}
  }
\end{table}
\

Table~\ref{tab:quality_benchmark_abalation} supplements the discussion in Section~\ref{subsec: ablation study}. It illustrates how the VBench score changes with different hyperparameters, including \textit{Shift Frequency} and \textit{Loop Step}. 

%%%%%%%%%%%%%%%%%%%%%%%%%%%%%%%%
\section{User Study Details}
\label{appendix: user study}

To complement the automated VBench metrics and specifically assess the necessity of the Stage-1 global guidance, we conducted a rigorous human evaluation. As noted in Section~\ref{subsec: ablation study}, automated metrics often fail to penalize structural artifacts such as object duplication or global incoherence in zero-shot tiled generation. This study specifically targets those perceptual aspects.

\subsection{Methodology}
We adopted a \textit{Two-Alternative Forced Choice (2AFC)} protocol, which is widely regarded as the gold standard for subjective video quality assessment. 

\noindent\textbf{Participants.} We recruited a group of $N=10$ evaluators. The group consisted of graduate students and researchers with background knowledge in computer vision and generative AI, ensuring they were sensitive to generation artifacts.

\noindent\textbf{Stimuli.} We randomly selected 10 prompts from the VBench dataset, covering various categories with dynamic scenes. For each prompt, we generated two video clips at 2K resolution:
1) Zero-Shot Generation: Direct generation using tile parallelism without Stage-1 low-resolution guidance.
2) \Mname{}: The full pipeline using Stage-1 global guidance followed by Stage-2 generation.
Both settings utilized the same random seed and CogVideoX model to ensure a fair comparison.

\noindent\textbf{Interface and Procedure.} The evaluation was conducted via a web-based interface. For each trial:
1) The two videos were displayed side-by-side.
2) The position (left vs. right) of the methods was randomized to prevent positional bias.
3) The text prompt was displayed above the videos.
4) The evaluators were blind to the method identifiers.

\subsection{Evaluation Criteria}
Participants were asked to select the video that was superior based on the following specific criteria, rather than just general aesthetic appeal:

\noindent \textbf{Global Structural Coherence.} Does the subject maintain a unified structure across the frame? 

\noindent \textbf{Semantic Fidelity.} Does the generated high-resolution video faithfully represent the prompt without adding hallucinatory objects?

\subsection{Results and Analysis}
We collected a total of $100$ pairwise comparisons. The results demonstrate a significant preference for the \Mname{} full pipeline. As shown in Table~\ref{tab:zero-shot}, users preferred \Mname{} in the vast majority of cases. 
% --- END OF APPENDIX CONFIGURATION ---

\end{document}